\pdfoutput=1
\documentclass{article}
\usepackage{iclr2027_conference,times}
\iclrfinalcopy
\usepackage[T1]{fontenc}
\usepackage[utf8]{inputenc}
\usepackage{graphicx,booktabs,amsmath,amssymb,array,multirow}
\usepackage{xcolor}
\usepackage{tikz}
\usetikzlibrary{arrows.meta,positioning,fit,backgrounds,calc,shapes.misc}
\usepackage{hyperref,url}
\hypersetup{colorlinks=true,linkcolor=blue!60!black,citecolor=blue!60!black,urlcolor=blue!60!black,
  pdftitle={Beyond Lip Sync: Reference-Grounded Oral Refinement for Audio-Driven Portrait Animation},
  pdfauthor={Bangxun Tang}}
\usepackage{microtype}
\usepackage{float}
\usepackage{animate}
\newcommand{\gen}{\hat{x}}
\newcommand{\refc}{r}
\newcommand{\crop}{\mathcal{C}}
\newcommand{\judge}{D_\psi}
\newcommand{\enc}{\phi_\psi}
\newcommand{\parser}{\Psi}
\newcommand{\ours}{\textsc{RGOR}}

\title{Beyond Lip Sync:\\Reference-Grounded Oral Refinement\\for Audio-Driven Portrait Animation}
\author{Bangxun Tang \\
University of California, Irvine \\
\texttt{bangxunt@uci.edu}}
\begin{document}
\maketitle
\lhead{Preprint. Under review.}
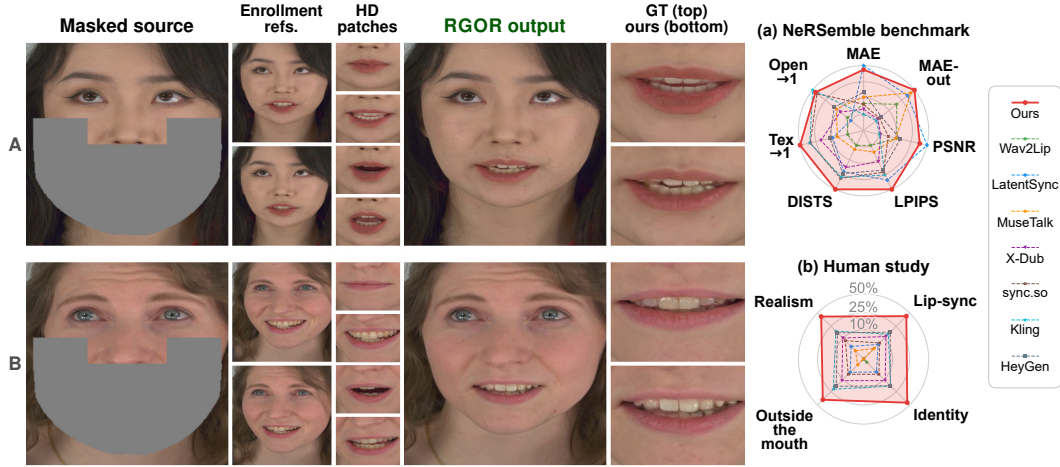
\begin{figure}[H]
\centering
\begin{minipage}[c]{97.5mm}
\begin{tikzpicture}[x=1mm,y=1mm,
  colhead/.style={font=\sffamily\bfseries\scriptsize,anchor=base,inner sep=0pt},
  rowlab/.style={font=\sffamily\bfseries\scriptsize,text=black!70,inner sep=0pt,anchor=center},
]
\useasboundingbox (0,-56.949) rectangle (97.5,4.449);
\node[colhead] at (15.87,0.9) {Masked source};
\node[colhead,font=\sffamily\bfseries\tiny] at (36.31,3.017) {Enrollment};
\node[colhead,font=\sffamily\bfseries\tiny] at (36.31,0.9) {refs.};
\node[colhead,font=\sffamily\bfseries\tiny] at (47.69,3.017) {HD};
\node[colhead,font=\sffamily\bfseries\tiny] at (47.69,0.9) {patches};
\node[colhead,text=green!35!black] at (65.81,0.9) {\ours{} output};
\node[colhead,font=\sffamily\bfseries\tiny] at (88.63,3.017) {GT (top)};
\node[colhead,font=\sffamily\bfseries\tiny] at (88.63,0.9) {ours (bottom)};
\node[rowlab] at (1.1,-14.01) {A};
\node[rowlab] at (1.1,-42.94) {B};
\node[inner sep=0pt,anchor=north west] at (2.5,0) {\animategraphics[loop,width=95mm]{2}{figures/teaser/anim/frame_}{00}{05}};
\end{tikzpicture}
\end{minipage}\hfill
\begin{minipage}[c]{30mm}
\centering
\includegraphics{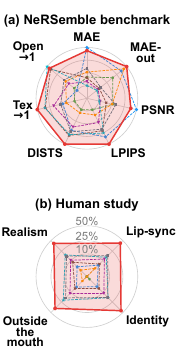}%
\end{minipage}\hfill
\begin{minipage}[c]{10mm}
\centering
\includegraphics{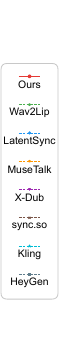}%
\end{minipage}%
\caption{Given a video with the mouth masked out, its speech, and enrollment frames and HD patches of the
same person, \ours{} renders that person's own lips and teeth in sync with the audio and leaves the rest of
the face untouched. Readers can click and play the video clips in this figure using \textcolor{red}{Adobe Acrobat}.}
\label{fig:teaser}
\end{figure}

\begin{abstract}
We present \ours{} (Reference-Grounded Oral Refinement), an audio-driven lip-sync
framework that renders the mouth of the specific person being dubbed rather than a
generic one. Existing lip-sync systems follow the audio closely and keep the face
recognizable, yet the mouth they render is an average mouth: the shape and texture of
the lips, the arrangement of the teeth, and how much of them shows as the mouth opens
are not that person's. The problem persists because nothing in current training or
evaluation asks for the person's own mouth: perceptual losses accept any plausible
mouth, face identity is carried mostly by the skin around it, and the released inference code of inpainting systems uses the unmasked target frame
as the reference, which hides the gap. To
address this, \ours{} conditions every generated frame on frames from separate
enrollment recordings of the same person and on HD patches of the mouth that bypass the
VAE, and trains the generator against a paired
judge that compares each rendered mouth with the person's reference and learns to
reject a realistic mouth of someone else. We further build an evaluation protocol and use
it to compare open-source and commercial lip-sync systems on held-out identities. Experiments show that \ours{}
achieves the best or second-best result on most metrics, and preserves the person's
own lip and dental detail while keeping synchronization and the rest of the face intact.

\end{abstract}

\section{Introduction}
Audio-driven lip synchronization has advanced quickly: current systems follow the audio
closely, render sharp frames, and keep the face recognizable to a face-recognition
network \citep{p_wav2lip,li2024latentsync,zhang2025musetalk}. Yet a viewer who knows
the person notices at once that the mouth is not theirs: the lips have a smooth,
generic texture, the teeth form a neat, even row, and how much of them shows as the
mouth opens follows an average face rather than this one. The mouth is a small part of
the frame, but it is where speech is seen, and replacing it with an average mouth
is exactly the change a dub should not make.

This failure goes unnoticed because neither training nor evaluation asks for the
person's own mouth. Perceptual losses reward any mouth with realistic statistics;
face-identity embeddings are dominated by the skin and face shape around the mouth;
and the released inference pipelines of LatentSync \citep{li2024latentsync}, MuseTalk
\citep{zhang2025musetalk}, and Wav2Lip \citep{p_wav2lip} use the unmasked frame being edited as the reference, so a paired test shows the model the
very mouth it must produce, a leakage the Wav2Lip authors already noted. Given a reference
from a different recording of the same person instead, the mouth fidelity of LatentSync and
Wav2Lip drops markedly (Appendix~\ref{app:leak}).

We propose Reference-Grounded Oral Refinement (\ours{}), which builds on LatentSync
and changes what its generator is conditioned on and trained toward. First, each frame
of a training window receives its own reference frame from separate enrollment
recordings of the same person, chosen so that the teeth are often visible; these
scattered references show the person's mouth in many shapes but offer no motion to copy. Second, since the
VAE of the latent generator compresses away the fine structure of the teeth and lips,
HD patches of the person's mouth reach the generator in pixel space and bypass the VAE.
Third, an available reference is not necessarily a used one, so we train, together with the
generator, a reference-contrastive paired judge that compares the rendered mouth with the person's
reference and rejects a realistic mouth of someone else, which a perceptual loss cannot do.
Fourth, since the judge compares appearance rather than
geometry, a shape term compares the mouth opening and lip regions of the output with
those of the real frame, as segmented by a frozen face parser. The four parts act as one:
the scattered references keep the generator from copying a mouth motion, the HD patches
carry the person's detail past the VAE, the judge steers the generator toward this
person's mouth, and the shape term keeps the opening in step with the speech.

To evaluate this, we hold out a set of identities of
NeRSemble \citep{kirschstein2023nersemble} from all training and split each person's
recordings into enrollment clips, which a system may use, and query clips, which it
must reproduce. Every system receives the query with the mouth removed or, if it cannot
take a masked input, a different recording of the person with the query audio. Our
contributions are summarized as follows:
\begin{itemize}\setlength{\itemsep}{1pt}
\item We propose \ours{}, a reference-grounded framework for audio-driven lip sync that
renders the specific person's lips and teeth rather than a generic mouth, by
conditioning each frame on scattered enrollment references of that person and on HD
patches of the mouth that bypass the VAE.
\item We formulate oral identity as a paired comparison between the rendered mouth and
the person's own reference, trained with a reference-contrastive judge whose negatives
include real mouths of other people, and complement it with a shape term that keeps the
mouth opening and the lip shape faithful to the real frame.
\item We introduce an evaluation protocol that withholds the target mouth from every
system that accepts a separate reference. It measures how much the same-frame reference of released lip-sync inpainters
inflates paired reconstruction scores, and under it we compare \ours{} with open-source and commercial lip-sync
systems on identities never seen in training.
\end{itemize}

\section{Related Work}
\paragraph{Audio-driven visual dubbing.}
Audio-driven visual dubbing re-renders the mouth of a talking-face video so that it
follows a new audio track. Wav2Lip \citep{p_wav2lip} framed the task as inpainting
supervised by a pretrained synchronization expert \citep{chung2016syncnet}, and later
GAN systems strengthened this formulation with lip-reading, expression, deformation,
and landmark priors \citep{p_talklip,p_vretalk,p_dinet,p_iplap}. Diffusion models
carried the same formulation forward, in latent and pixel space
\citep{p_difftalk,p_diff2lip} and with SyncNet supervision on decoded frames
\citep{li2024latentsync}, while MuseTalk reaches real time with single-step latent
inpainting \citep{zhang2025musetalk}. A parallel line animates the
whole portrait from a single image \citep{p_hallo,p_echo,p_omnihuman}, and recent
systems explore mask-free editing that replaces inpainting, released on the Wan2.2
video model \citep{he2025xdub,p_wan}, production-oriented post-training
\citep{li2026tbdub}, and commercial services that take a video and an audio track with
no reference \citep{p_syncso,p_kling,p_heygen}. Despite this progress, the inpainting
methods condition on a reference frame of the speaker but leave whose mouth is rendered
unsupervised, so the lips and teeth drift toward an average mouth. \ours{} builds on
LatentSync, keeps its inpainting formulation, and adds this supervision with
references from separate recordings of the person.

\paragraph{Identity and texture preservation with references.}
Generative models commonly preserve the identity and texture of a specific subject by
conditioning on reference images. Appearance features of the reference can be injected
through attention \citep{p_animateanyone}, identity embeddings through decoupled cross-attention \citep{p_ipadapter,p_instantid}
or merged into the text embedding \citep{p_photomaker}, or the subject fitted
into the weights themselves \citep{p_dreambooth,p_lora}. Reference-based face
restoration and super-resolution condition the restorer on clean reference images, of
the same person in the face-restoration case
\citep{p_refldm,p_faceme,p_instantrestore,p_ttsr,p_c2matching}. In lip sync, references
enter through an identity perceiver that counters averaged lips \citep{zhu2025unavglip}
or through reference textures warped by motion fields \citep{p_iptalker,p_efficientsync}.
These mechanisms make identity available to the generator, yet nothing in their
objectives requires the output to use it, so a reference branch can be read for
lighting and mouth opening while ignoring who the person is. \ours{} pairs its references with an objective that compares the
rendered mouth against them.

\paragraph{Adversarial and contrastive judges.}
Adversarial training supervises a generator with a learned discriminator
\citep{p_gan}; conditional GANs also show the discriminator the conditioning input
\citep{p_cgan,p_pix2pix}, and the projection discriminator scores an image embedding
against it \citep{p_projdisc}, so the judge asks whether a pair is realistic given the
condition. Later discriminators are built on frozen pretrained features
\citep{p_projectedgan,p_visionaided,p_stylegant} and inherit their blind spots, and
general-purpose self-supervised features such as DINOv2 \citep{p_dinov2} are not
trained to tell one person's inner mouth from another's. In lip sync, the pairwise
SyncNet expert scores whether audio and video match and serves as both loss and metric
\citep{chung2016syncnet,p_wav2lip}, and as a loss it can be gamed by the generator.
Perceptual losses compare an output with its target in a deep feature space and
tolerate misalignment to varying degrees
\citep{p_perceptual,zhang2018lpips,ding2021dists,p_contextual}. Each of these judges
asks whether an output is realistic, synchronized, or close to its target, and none
tells one person's teeth from another's. The judge of \ours{} is closest to Siamese
verification \citep{p_chopra2005}: it scores whether two mouths belong to the same
person, and because real mouths of other people are negatives, realism alone cannot
satisfy it.

\paragraph{Evaluating lip-sync fidelity and identity.}
Lip-sync evaluation measures synchronization with LSE-C and LSE-D from SyncNet
\citep{chung2016syncnet,p_wav2lip}, identity with ArcFace cosine similarity
\citep{p_arcface}, and texture or distribution match with LPIPS, DISTS, and FVD
\citep{zhang2018lpips,ding2021dists,p_fvd}. Each has a blind spot
for the mouth of a specific person: LSE ignores whose mouth moves, ArcFace on a whole
face is dominated by the copied upper face, PSNR rewards blur, and LPIPS and DISTS
accept texture with the right statistics. Paired reconstruction adds a leak: the released LatentSync, MuseTalk, and
Wav2Lip inference code takes the reference from the very frame being reconstructed and
so shows the model the answer, a leakage the Wav2Lip authors noted and avoided by
evaluating with audio from another video \citep{p_wav2lip}. NeRSemble
\citep{kirschstein2023nersemble} records separate sentence and expression sequences
for each identity, which allows a clean split into enrollment and query clips, and on
this split we build a strict cross-clip protocol (Sec.~\ref{sec:protocol}).

\section{Method}\label{sec:method}

\subsection{Overview}\label{sec:overview}
\begin{figure}[t]
\centering
\definecolor{rgPanIn}{HTML}{FFF7E6}\definecolor{rgBrdIn}{HTML}{E5C67E}\definecolor{rgLabIn}{HTML}{9A6B10}
\definecolor{rgPanGe}{HTML}{ECF5EE}\definecolor{rgBrdGe}{HTML}{A5C9B0}\definecolor{rgLabGe}{HTML}{1E7A45}
\definecolor{rgPanRe}{HTML}{EBF2FB}\definecolor{rgBrdRe}{HTML}{A6C2E3}\definecolor{rgLabRe}{HTML}{1F5FA8}
\definecolor{rgPanOb}{HTML}{FCEEEF}\definecolor{rgBrdOb}{HTML}{DFACB0}\definecolor{rgLabOb}{HTML}{B4232A}
\definecolor{rgFrz}{HTML}{EFECF8}\definecolor{rgFrzB}{HTML}{9C93C4}
\definecolor{rgTrn}{HTML}{FBDED8}\definecolor{rgTrnB}{HTML}{C0392B}
\definecolor{rgOpp}{HTML}{E2EFF8}\definecolor{rgOppB}{HTML}{7FA9C8}
\definecolor{rgJdg}{HTML}{F8DCDF}
\definecolor{rgChZ}{HTML}{C9C9C9}\definecolor{rgChM}{HTML}{555555}
\definecolor{rgChC}{HTML}{F2D08A}\definecolor{rgChR}{HTML}{8FBDE6}
\definecolor{rgGrad}{HTML}{C0392B}\definecolor{rgPos}{HTML}{1E7A45}\definecolor{rgNeg}{HTML}{B4232A}
\newcommand{\rgs}[1]{{\fontsize{6.6}{7.4}\selectfont\color{black!55}#1}}
\newcommand{\rgm}[1]{{\fontsize{6.6}{7.4}\selectfont #1}}
\newcommand{\rgsnow}{\tikz[baseline=-2pt,x=1pt,y=1pt]{%
  \foreach \a in {30,90,150}{%
    \draw[rgFrzB!85!black,line width=0.34pt,line cap=round] (\a:3.3) -- (\a+180:3.3);}%
  \foreach \a in {30,90,150,210,270,330}{%
    \draw[rgFrzB!85!black,line width=0.34pt,line cap=round] (\a:1.85) -- ++(\a+42:1.25);%
    \draw[rgFrzB!85!black,line width=0.34pt,line cap=round] (\a:1.85) -- ++(\a-42:1.25);}%
}}
\resizebox{\linewidth}{!}{%
\begin{tikzpicture}[x=1mm,y=1mm,
  font=\sffamily\footnotesize,
  pan/.style={rounded corners=3mm,line width=0.6pt,inner sep=0pt},
  pantit/.style={font=\sffamily\bfseries\small,anchor=north west,inner sep=0pt},
  blk/.style={draw=black!55,line width=0.5pt,rounded corners=1.4mm,fill=white,
              align=center,inner xsep=1.4mm,inner ysep=1.1mm},
  frz/.style={blk,fill=rgFrz,draw=rgFrzB},
  trn/.style={blk,fill=rgTrn,draw=rgTrnB,line width=0.7pt},
  opn/.style={blk,fill=rgOpp,draw=rgOppB},
  jdg/.style={blk,fill=rgJdg,draw=rgLabOb,line width=0.7pt},
  img/.style={inner sep=0pt,draw=black!45,line width=0.35pt},
  nte/.style={font=\sffamily\fontsize{6.6}{7.9}\selectfont,text=black!58,align=center,inner sep=0pt},
  lbl/.style={font=\sffamily\fontsize{6.6}{7.4}\selectfont,align=center,inner sep=0pt},
  onl/.style={font=\sffamily\fontsize{6.6}{7.4}\selectfont,fill=white,inner sep=0.7pt,align=center},
  arr/.style={-{Latex[length=1.9mm,width=1.5mm]},line width=0.65pt,draw=black!72,
              rounded corners=1.4mm},
  arrb/.style={-{Latex[length=1.6mm,width=1.3mm]},line width=0.5pt,draw=black!60,
               rounded corners=1.2mm},
  bp/.style={-{Latex[length=1.9mm,width=1.5mm]},line width=0.65pt,draw=rgGrad,
             dash pattern=on 1.5mm off 0.9mm,rounded corners=1.4mm},
]
\useasboundingbox (0,-0.5) rectangle (152,80);

\begin{scope}[on background layer]
  \node[pan,fill=rgPanIn,draw=rgBrdIn,fit={(1.5,46)(40,78)}]     (pIn)  {};
  \node[pan,fill=rgPanGe,draw=rgBrdGe,fit={(43,46)(150.5,78)}]   (pGen) {};
  \node[pan,fill=rgPanRe,draw=rgBrdRe,fit={(1.5,0.5)(78,43)}]    (pRef) {};
  \node[pan,fill=rgPanOb,draw=rgBrdOb,fit={(81,0.5)(150.5,43)}]  (pObj) {};
\end{scope}
\node[pantit,text=rgLabIn] at (4.2,76.8) {INPUTS};
\node[pantit,text=rgLabGe] at (45.5,76.8)
  {GENERATOR\hspace{3.5pt}\rgs{LatentSync-1.6 parent, inpainting UNet}};
\node[pantit,text=rgLabRe] at (4.2,41.6) {ENROLLMENT BANK\hspace{3.5pt}\rgs{(never a query)}};
\node[pantit,text=rgLabOb] at (83.5,41.6) {OBJECTIVES};

\node[blk,minimum width=36mm,minimum height=6mm,anchor=center] (bxA) at (20.75,69.2) {};
\foreach \x/\h in {-3.5/1.2,-3.0/2.6,-2.5/1.7,-2.0/3.6,-1.5/2.2,-1.0/3.2,-0.5/1.4,
                   0/2.8,0.5/1.3,1.0/3.4,1.5/1.8,2.0/2.4,2.5/1.1,3.0/2.9,3.5/1.6}
  \draw[line width=0.55pt,line cap=round,draw=rgLabIn]
        (7.95+\x,69.2-\h/2) -- (7.95+\x,69.2+\h/2);
\node[anchor=west,inner sep=0pt] at (13.2,69.2) {\textbf{TARGET AUDIO}};

\node[blk,minimum width=36mm,minimum height=9.5mm,anchor=center] (bxB) at (20.75,58.8) {};
\node[img] at (7.95,58.8) {\includegraphics[width=7mm]{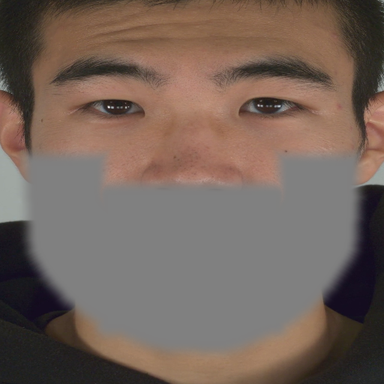}};
\node[anchor=west,inner sep=0pt,align=left] at (13.2,58.8) {\textbf{MASKED}\\[-0.3ex]\textbf{SOURCE} $S$};

\node[nte,anchor=north] at (20.75,53.2)
     {one window: $F{=}16$ aligned\\$512{\times}512$ frames at 25\,fps};

\node[frz,minimum width=19.5mm,minimum height=7mm] (whis) at (54.8,69.2)
     {\textbf{WHISPER}\\[-0.2ex]\rgsnow};
\node[frz,minimum width=19.5mm,minimum height=9mm] (venc) at (54.8,58.8)
     {\textbf{VAE ENC} $\mathcal{E}$\\[-0.2ex]\rgsnow};

\begin{scope}[shift={(67.9,53.15)}]
  \def\cW{5.8}\def\cH{8.6}\def\dX{0.4385}\def\dY{0.33}
  \fill[rgChZ] (0,0) rectangle (\cW,\cH);                %
  \foreach \k/\cl in {0/rgChZ,1/rgChZ,2/rgChZ,3/rgChZ,4/rgChM,
                      5/rgChC,6/rgChC,7/rgChC,8/rgChC,
                      9/rgChR,10/rgChR,11/rgChR,12/rgChR}{
    \fill[\cl!92!white] (\k*\dX,\cH+\k*\dY) -- (\cW+\k*\dX,\cH+\k*\dY)
        -- (\cW+\k*\dX+\dX,\cH+\k*\dY+\dY) -- (\k*\dX+\dX,\cH+\k*\dY+\dY) -- cycle;
    \fill[\cl!72!black] (\cW+\k*\dX,\k*\dY) -- (\cW+\k*\dX,\cH+\k*\dY)
        -- (\cW+\k*\dX+\dX,\cH+\k*\dY+\dY) -- (\cW+\k*\dX+\dX,\k*\dY+\dY) -- cycle;
    \draw[black!45,line width=0.25pt]
        (\k*\dX+\dX,\cH+\k*\dY+\dY) -- (\cW+\k*\dX+\dX,\cH+\k*\dY+\dY)
        -- (\cW+\k*\dX+\dX,\k*\dY+\dY);}
  \draw[black!55,line width=0.4pt] (0,0) rectangle (\cW,\cH);
  \draw[black!55,line width=0.4pt] (\cW,0) -- (\cW+13*\dX,13*\dY)
      -- (\cW+13*\dX,\cH+13*\dY) -- (13*\dX,\cH+13*\dY) -- (0,\cH);
\end{scope}
\coordinate (stkW) at (67.9,58.8);
\coordinate (stkE) at (79.4,58.8);
\node[nte,anchor=north] at (73.65,52.35)
     {13 channels\\$z_t\,{\mid}\,$mask$\,{\mid}\,\mathcal{E}S\,{\mid}\,\mathcal{E}\refc$};

\node[trn,minimum width=20mm,minimum height=15mm] (unet) at (92.75,58.8)
     {\textbf{3D UNET} $\epsilon_\theta$\\[-0.2ex]\rgs{trainable}\\[0.3ex]
      \rgm{audio cross-attn}\\[-0.2ex]\rgm{estimate $\hat z_0$}};
\node[frz,minimum width=15mm,minimum height=9mm] (vdec) at (116.15,58.8)
     {\textbf{VAE DEC} $\mathcal{D}$\\[-0.2ex]\rgsnow};
\node[opn,minimum width=19mm,minimum height=9mm] (comp) at (139,58.8)
     {\textbf{COMPOSE}\\[-0.2ex]\rgm{$(1{-}\alpha)S{+}\alpha\gen$}};

\foreach \i/\x/\y in {1/11.2/25.8, 2/26.1/25.8, 3/41.0/25.8,
                      4/11.2/10.9, 5/26.1/10.9, 6/41.0/10.9}
  \node[img] at (\x,\y) {\includegraphics[width=14mm]{figures/pipeline/ref\i.png}};

\node[nte,anchor=south,text=rgLabRe] at (62,33.9) {teeth-visibility prior};
\node[opn,minimum width=19mm,minimum height=12mm] (scat) at (62,26.8)
     {\textbf{SCATTER}\\[-0.2ex]\rgm{one reference}\\[-0.4ex]\rgm{per output frame}};

\node[img] at (64.7,10.9)      {\includegraphics[width=11mm]{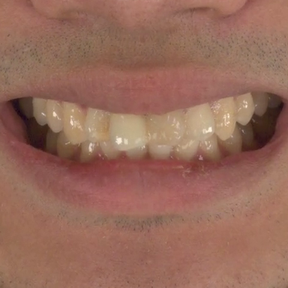}};
\node[img] at (64.1,11.45)     {\includegraphics[width=11mm]{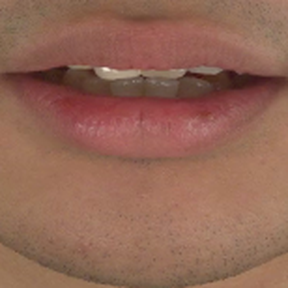}};
\node[img] (rstk) at (63.5,12) {\includegraphics[width=11mm]{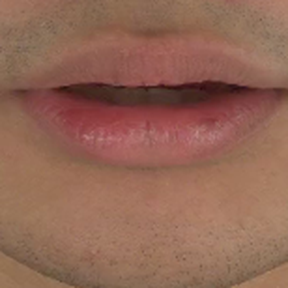}};
\node[nte,anchor=north] at (64.1,4.8) {$\refc_1\ldots\refc_{16}$};

\node[img] (oGt)   at (121.5,35.7) {\includegraphics[width=11mm]{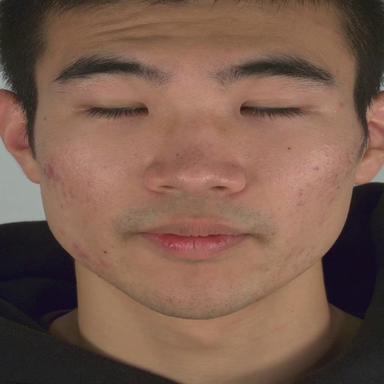}};
\node[img] (oOurs) at (139,35.7)   {\includegraphics[width=11mm]{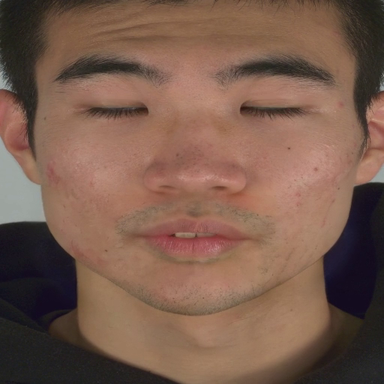}};
\node[lbl,text=black!60] at (130.25,35.7) {vs.};
\node[lbl,anchor=north,text=black!60] at (121.5,29.6) {GT $y$};
\node[lbl,anchor=north,text=black!60] at (139,29.6)   {ours $\tilde x$};

\node[blk,inner ysep=0.7mm,minimum width=28mm] (pix) at (97.5,34.4)
     {\textbf{IMAGE TERMS}\\[-0.2ex]\rgs{LPIPS $\cdot$ DISTS $\cdot$ boundary}};
\node[blk,inner ysep=0.7mm,minimum width=28mm] (shp) at (97.5,26.7)
     {\textbf{SHAPE TERM}\\[-0.2ex]\rgs{parser $\parser$: opening, lips}\,\rgsnow};
\draw[arrb,-] (oGt.west) -- (114.4,35.7) -- (114.4,26.7);
\draw[arrb] (114.4,34.4) -- (pix.east);
\draw[arrb] (114.4,26.7) -- (shp.east);

\node[img] (jR) at (88.5,15.25)  {\includegraphics[width=9mm]{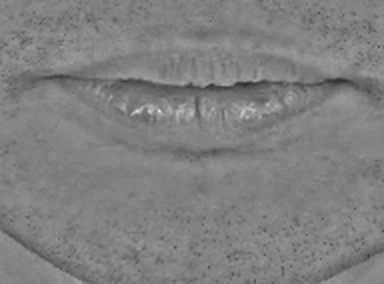}};
\node[img] (jG) at (99,15.25)    {\includegraphics[width=9mm]{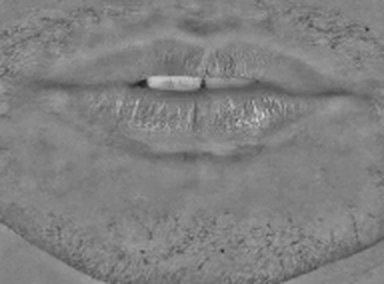}};
\node[img] (jO) at (109.5,15.25) {\includegraphics[width=9mm]{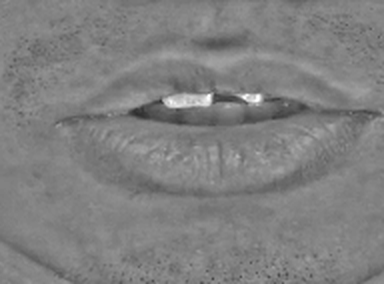}};
\node[lbl,text=rgPos,anchor=south] at (88.5,19.9)  {real $y$};
\node[lbl,text=rgNeg,anchor=south] at (99,19.9)    {gen $\tilde x$};
\node[lbl,text=rgNeg,anchor=south] at (109.5,19.9) {other $y'$};
\node[img] (jRef) at (122.5,15.25) {\includegraphics[width=9mm]{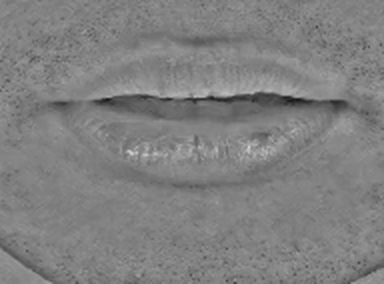}};
\node[lbl,anchor=west,align=left,text=black!60] at (128.5,16.5)
     {reference $\refc_{j(f)}$,\\one the UNet saw};
\node[nte,anchor=west,align=left] at (128.5,9.4)
     {grey high-passed\\oral box $\crop$};
\node[jdg,inner ysep=0.7mm,minimum width=30mm] (judge) at (99,4.9)
     {\textbf{PAIRED JUDGE} $\judge$\\[-0.2ex]\rgs{same person as $\refc$?}};

\draw[arr] (bxA.east) -- (whis.west);
\draw[arr] (bxB.east) -- (venc.west);
\draw[arr] (whis.east) -- (92.75,69.2) -- (unet.north);
\node[nte,anchor=south] at (77.5,69.8) {cross-attention};
\node[img] at (121.1,68.9)       {\includegraphics[width=8mm]{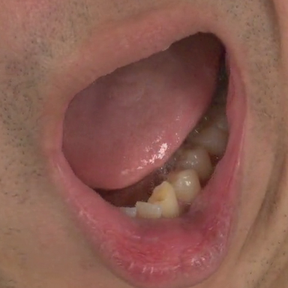}};
\node[img] at (120.6,69.4)       {\includegraphics[width=8mm]{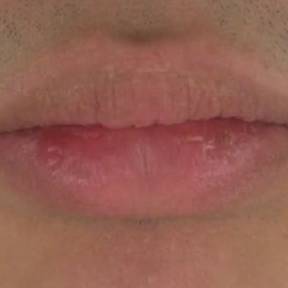}};
\node[img] (hdstk) at (120.1,69.9) {\includegraphics[width=8mm]{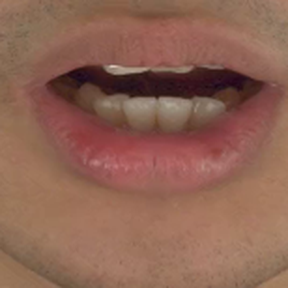}};
\node[lbl,anchor=west,align=left,text=black!60] at (126.6,69.4) {4 HD patches\\(bypass the VAE)};
\draw[arr] ([yshift=-0.7mm]hdstk.west) -- (98.75,69.2) -- ([xshift=6mm]unet.north);
\node[nte,anchor=south] at (107.4,69.8) {gated attn.};
\draw[arr] (venc.east) -- (stkW);
\draw[arr] (stkE) -- (unet.west);
\draw[arr] (unet.east) -- (vdec.west);
\draw[arr] (vdec.east) -- (comp.west);
\draw[arr] (48.8,26.8) -- (scat.west);
\draw[arr] (scat.south) -- (62,17.8);
\draw[arr] (70.7,12) -- (75.8,12) -- (75.8,44.5) -- (54.8,44.5) -- (venc.south);
\node[onl] at (61.5,44.5) {$\refc_f$};
\draw[arr] (comp.south) -- (139,41.5);
\node[onl] at (139,44.5) {$\tilde x$};
\draw[arrb] (jR.south) -- (88.5,8.45);
\draw[arrb] (jG.south) -- (99,8.45);
\draw[arrb] (jO.south) -- (109.5,8.45);
\node[onl,inner sep=0.3pt,text=rgPos] at (88.5,10.9) {$+$};
\node[onl,inner sep=0.3pt,text=rgNeg] at (99,10.9)   {$-$};
\node[onl,inner sep=0.3pt,text=rgNeg] at (109.5,10.9){$-$};
\draw[arrb] (jRef.south) -- (122.5,4.9) -- (judge.east);
\draw[bp] ([xshift=-3.5mm]oOurs.north) -- (135.5,48) -- (92.75,48) -- (unet.south);   %
\node[nte,anchor=south,text=rgGrad] at (114,48.6)
     {$\partial\mathcal{L}/\partial\theta$ through the composite and $\mathcal{D}$};
\end{tikzpicture}}
\caption{Overview of \ours{}. Each output frame gets its own scattered enrollment reference, and four HD patches bypass the VAE through gated attention; the masked source, the references, and the audio drive the inpainting UNet, and the composited output is trained with image terms, a shape term, and a paired judge that asks whether the rendered mouth belongs to the referenced person.}
\label{fig:pipeline}
\end{figure}

\ours{} builds on LatentSync-1.6 \citep{li2024latentsync}, which we call the parent: a
3D UNet $\epsilon_\theta$ that denoises windows of $F{=}16$ aligned $512{\times}512$
face frames in the latent space of a frozen VAE $(\mathcal{E},\mathcal{D})$,
conditioned on Whisper audio features \citep{p_whisper} through cross-attention. For
each output frame it receives the noisy latent, the lower-face mask, the masked source
frame $S_f$, which carries the pose and the surrounding skin, and one reference frame
$\refc_f$. Since the lower face of the source is masked, the references are the only
inputs that show the person's mouth. During training, we noise the ground-truth latent at a
random timestep and convert the model's noise prediction into an estimate $\hat z_0$ of
the clean latent, which is decoded and composited back into the source through the
feathered mask $\alpha\in[0,1]^{512\times512}$ of the editable region:
\begin{equation}
\gen=\mathcal{D}(\hat z_0),\qquad
\tilde x_f=(1-\alpha)\odot S_f+\alpha\odot\gen_f,\qquad f=1,\dots,F .
\label{eq:composite}
\end{equation}
All training objectives act on the composite $\tilde x$, with gradients through the
frozen decoder. \ours{} changes four things in the parent (Fig.~\ref{fig:pipeline}):
the references it is conditioned on (Sec.~\ref{sec:scatter}), HD patches of the mouth
that bypass the VAE (Sec.~\ref{sec:hdpatch}), a paired judge that compares the rendered
mouth with those references (Sec.~\ref{sec:judge}), and a shape term on the mouth
opening (Sec.~\ref{sec:shape}).

\subsection{Scattered enrollment references}\label{sec:scatter}
Each person has an enrollment bank: aligned frames of a few other clips of that
person that never serve as queries, each scored by a frozen face parser for
visible-teeth area. Instead of a frame of the clip being edited, each of the sixteen
frames of a training window receives its own reference drawn from the bank, with a
preference for frames that show teeth. The references are unrelated in time, so
together they show the person's lips and teeth in many mouth shapes without a motion the
model could copy. They enter the UNet through its reference input channels. At
inference, sixteen bank frames are picked in a fixed order. We also explored a
contiguous window of enrollment frames and a single static reference; the first carries
the enrollment clip's mouth motion into the output and the second shows only one mouth
state, so we use scattered references.

\subsection{HD patches}\label{sec:hdpatch}
The references reach the UNet through the VAE, which compresses each frame eight-fold
along each side; at this scale a tooth spans only a few latent cells, and the gaps
between the teeth and the creases of the lips are lost before the generator sees them.
\ours{} therefore adds HD patches: four full-resolution crops of the person's mouth from
the enrollment bank, covering its states from closed to widest open, including the frame
with the most visible teeth. The HD patches bypass the VAE. A reference encoder encodes
them directly from their pixels, and gated attention adapters on the up blocks of the UNet let every
position of the output attend to them, so the fine structure of the person's lips and
teeth reaches the generator at the resolution of the video.

\subsection{Reference-contrastive paired judge}\label{sec:judge}
\begin{figure}[t]
\centering
\definecolor{fdCap}{HTML}{FFF5D6}\definecolor{fdCapL}{HTML}{8F6200}
\definecolor{fdGen}{HTML}{E4F5EA}\definecolor{fdGenL}{HTML}{1E7A45}
\definecolor{fdRef}{HTML}{E3EFFB}\definecolor{fdRefL}{HTML}{1F5FA8}
\definecolor{fdHeld}{HTML}{EFE9F8}\definecolor{fdHeldL}{HTML}{5A3E9E}
\definecolor{fdBar}{HTML}{F1F1F4}
\resizebox{\linewidth}{!}{%
\begin{tikzpicture}[x=1mm,y=1mm,
  font=\sffamily\scriptsize,
  panel/.style={draw=black!20,line width=0.4pt,rounded corners=1.8mm,inner sep=0pt},
  ttl/.style={font=\sffamily\bfseries\footnotesize,anchor=north west,inner sep=0pt},
  note/.style={font=\sffamily\scriptsize,text=black!60,align=center,anchor=north,inner sep=0pt},
  noteS/.style={font=\sffamily\scriptsize,text=black!60,align=center,anchor=south,inner sep=0pt},
  img/.style={inner sep=0pt,draw=black!55,line width=0.3pt},
  card/.style={inner sep=0.3mm,fill=white,draw=black!55,line width=0.3pt},
  tag/.style={font=\sffamily\scriptsize,inner sep=0.55pt,fill=white,fill opacity=0.86,text opacity=1,
              rounded corners=0.3mm,text=black!75},
  arr/.style={-{Latex[length=1.8mm,width=1.5mm]},line width=0.6pt,draw=black!70,rounded corners=1mm},
  hair/.style={line width=0.25pt,draw=black!35},
]
\useasboundingbox (0,0) rectangle (140,49.7);

\begin{scope}[on background layer]
  \node[panel,fill=fdCap,  fit={(0.8,31.2)(28.3,49.4)}]   (pCap)  {};
  \node[panel,fill=fdCap,  fit={(31.3,31.2)(54.8,49.4)}]  (pAlg)  {};
  \node[panel,fill=fdGen,  fit={(57.8,31.2)(86.8,49.4)}]  (pMsk)  {};
  \node[panel,fill=fdGen,  fit={(89.8,31.2)(139.2,49.4)}] (pWin)  {};
  \node[panel,fill=fdBar,draw=black!25,rounded corners=1.5mm,fit={(0.8,22.8)(139.2,29.4)}] (pBar) {};
  \node[panel,fill=fdHeld, fit={(0.8,0.3)(55.0,20.8)}]    (pHeld) {};
  \node[panel,fill=fdRef,  fit={(58.0,0.3)(139.2,20.8)}]  (pBank) {};
\end{scope}

\node[ttl,text=fdCapL] at (3.4,48.9) {CAPTURE};
\foreach \i in {-1,0,1}{\foreach \j in {-1,0,1}{%
  \pgfmathsetmacro{\cx}{9.0+\i*2.15}\pgfmathsetmacro{\cy}{40.2+\j*2.15}
  \pgfmathtruncatemacro{\ctr}{ifthenelse(\i==0 && \j==0,1,0)}
  \ifnum\ctr=1
    \draw[fill=fdCapL,draw=fdCapL,rounded corners=0.22mm] (\cx-0.9,\cy-0.76) rectangle (\cx+0.9,\cy+0.76);
    \fill[white] (\cx,\cy) circle (0.4);
  \else
    \draw[fill=white,draw=black!50,line width=0.25pt,rounded corners=0.22mm]
      (\cx-0.8,\cy-0.68) rectangle (\cx+0.8,\cy+0.68);
    \fill[black!45] (\cx,\cy) circle (0.32);
  \fi}}
\draw[arr,draw=fdCapL] (12.6,40.2) -- (15.3,40.2);
\node[card] at (21.6,41.6) {\includegraphics[width=7.8mm]{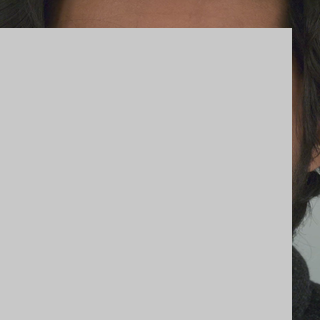}};
\node[card] at (20.6,40.6) {\includegraphics[width=7.8mm]{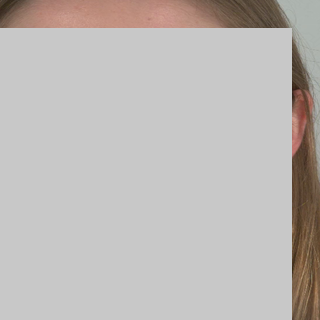}};
\node[card] at (19.6,39.6) {\includegraphics[width=7.8mm]{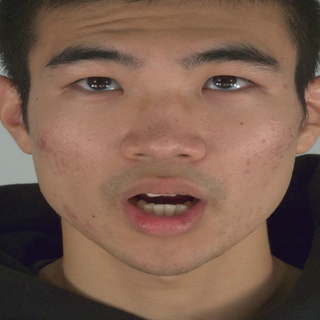}};

\node[ttl,text=fdCapL] at (33.9,48.9) {ALIGN};
\node[img] (algim) at (43.05,40.2) {\includegraphics[width=11.0mm]{figures/data/align_face.png}};
\foreach \sx/\sy in {-1/-1,-1/1,1/-1,1/1}{%
  \draw[line width=0.5pt,draw=fdCapL]
    (43.05+\sx*5.25,40.2+\sy*5.25) -- (43.05+\sx*3.5,40.2+\sy*5.25)
    (43.05+\sx*5.25,40.2+\sy*5.25) -- (43.05+\sx*5.25,40.2+\sy*3.5);}
\node[tag,anchor=south] at ([yshift=0.4mm]algim.south) {$512{\times}512$};

\node[ttl,text=fdGenL] at (60.4,48.9) {MASK\,{+}\,AUDIO};
\node[img] (mskim) at (64.5,40.2) {\includegraphics[width=11.0mm]{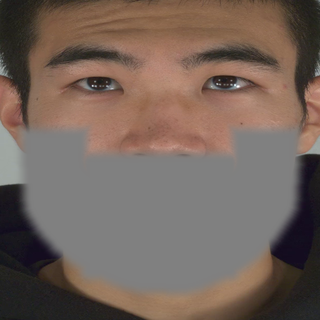}};
\draw[fill=white,draw=black!55,line width=0.3pt,rounded corners=0.5mm] (73.0,34.7) rectangle (84.2,45.7);
\foreach \i in {0,...,19}{%
  \pgfmathsetmacro{\hh}{0.5+3.4*abs(sin(\i*47+13))}
  \pgfmathsetmacro{\xx}{73.75+\i*0.53}
  \draw[fdRefL,line width=0.4pt,line cap=round] (\xx,42.2-\hh/2) -- (\xx,42.2+\hh/2);}
\draw[hair] (73.6,39.55) -- (83.6,39.55);
\foreach \i in {0,...,12}{\foreach \j in {0,...,4}{%
  \pgfmathsetmacro{\vv}{0.12+0.82*abs(sin(\i*57+\j*103+31))}
  \pgfmathsetmacro{\xx}{73.6+\i*0.78}
  \pgfmathsetmacro{\yy}{35.7+\j*0.64}
  \fill[fdRefL,opacity=\vv] (\xx,\yy) rectangle ++(0.68,0.52);}}
\node[noteS] at (72.3,31.8) {frozen Whisper features};

\node[ttl,text=fdGenL] at (92.4,48.9) {16-FRAME WINDOWS};
\fill[black!70,rounded corners=0.4mm] (90.25,36.45) rectangle (138.75,43.95);
\foreach \i in {0,...,21}{%
  \fill[white,rounded corners=0.12mm] (90.9+\i*2.2,43.15) rectangle ++(0.8,0.45);
  \fill[white,rounded corners=0.12mm] (90.9+\i*2.2,36.8) rectangle ++(0.8,0.45);}
\foreach \i/\f in {0/00,1/01,2/02,3/03,4/04,5/05,6/06,7/07}{%
  \node[inner sep=0pt] at (93.5+\i*6.0,40.2) {\includegraphics[width=6.0mm]{figures/data/win_\f.png}};}
\draw[line width=0.45pt,draw=black!55] (90.5,36.2) -- (90.5,35.8) -- (138.5,35.8) -- (138.5,36.2);
\node[noteS] at (114.5,31.8) {every 2nd frame of one window};

\node[inner sep=0pt] at (70,26.1) {\sffamily\scriptsize\setlength{\tabcolsep}{0pt}%
  \begin{tabular}{@{}l@{\hspace{4mm}}l@{}}
    \textcolor{black!75}{\bfseries CORPUS} &
      \textcolor{black!60}{2{,}900 clips (10 sentences per identity)\hspace{3pt}$\cdot$\hspace{3pt}%
      about 18.6k 16-frame units} \\[-0.2mm]
    \textcolor{black!75}{\bfseries IDENTITIES} &
      \textcolor{black!60}{290 in the training corpus\hspace{3pt}$\cdot$\hspace{3pt}%
      \textcolor{fdHeldL}{\bfseries 50 more held out for the benchmark}} \\
  \end{tabular}};

\node[ttl,text=fdHeldL] at (3.4,20.3) {HELD-OUT BENCHMARK};
\node[img] at (14.65,11.2) {\includegraphics[width=11.5mm]{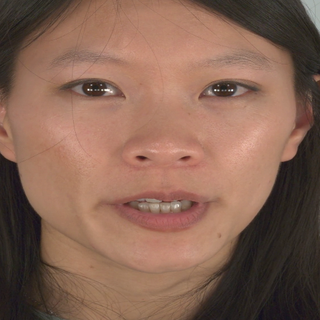}};
\node[img] at (27.65,11.2) {\includegraphics[width=11.5mm]{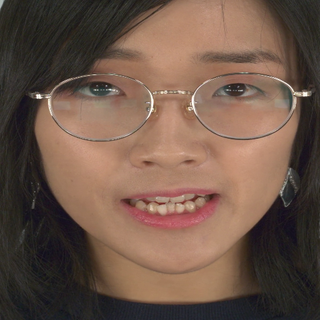}};
\draw[fill=white,draw=black!30,line width=0.3pt,rounded corners=0.5mm]
  (35.7,5.45) rectangle (47.2,16.95);
\draw[fill=white,draw=black!30,line width=0.3pt,rounded corners=0.5mm]
  (35.3,5.45) rectangle (46.8,16.95);
\draw[fill=black!6,draw=black!35,line width=0.3pt,rounded corners=0.5mm]
  (34.9,5.45) rectangle (46.4,16.95);
\node[font=\sffamily\bfseries\footnotesize,text=black!45] at (40.65,11.2) {$+48$};
\node[draw=fdHeldL,fill=fdHeldL,line width=0.4pt,rounded corners=0.7mm,inner xsep=1.4mm,inner ysep=0.85mm,
      font=\sffamily\bfseries\scriptsize,text=white] at (15.5,2.7) {NEVER TRAINED ON};
\node[anchor=west,font=\sffamily\scriptsize,text=black!60,inner sep=0pt] at (30.5,2.7) {100 query clips};

\node[ttl,text=fdRefL] at (60.6,20.3) {ENROLLMENT BANK};
\foreach \a/\b in {60.6/98.3,99.4/137.1}{%
  \draw[fill=black!11,draw=black!25,line width=0.25pt,rounded corners=0.35mm]
    (\a,15.8) rectangle (\b,17.0);}
\foreach \x in {60.78,63.37,67.07,72.98,79.27,86.66,92.94,99.92,105.80,112.72,115.14,118.60,124.13,130.01,133.47,136.93}{%
  \draw[line width=0.25pt,draw=black!45] (\x,16.05) -- (\x,16.75);}
\foreach \x in {67.44,78.53,96.64,107.18,129.32,136.58}{%
  \draw[fdRefL,line width=0.6pt,line cap=round] (\x,15.65) -- (\x,17.15);}
\foreach \i/\f in {0/1,1/2,2/3,3/4,4/5,5/6}{%
  \node[img] at (66.75+\i*12.84,10.3) {\includegraphics[width=12.3mm]{figures/data/bank_\f.png}};}
\node[noteS] at (98.9,2.0) {2 enrollment sentences per identity, never a query};

\draw[arr] (28.3,40.2) -- (31.3,40.2);
\draw[arr] (54.8,40.2) -- (57.8,40.2);
\draw[arr] (86.8,40.2) -- (89.8,40.2);
\draw[arr] (43.05,31.2) -- (43.05,29.5);
\draw[arr] (20.0,22.7) -- (20.0,20.8);
\draw[arr] (80.0,22.7) -- (80.0,20.8);
\end{tikzpicture}}
\caption{Data construction from NeRSemble.}
\label{fig:data}
\end{figure}

The paired judge $\judge$ decides whether a rendered mouth and a reference mouth belong
to the same person. A differentiable crop $\crop$ takes the fixed oral box of a frame,
converts it to grey, and high-passes it, so the judge reads the structure of the teeth
and lips, such as tooth boundaries, gaps, and lip creases, rather than their colour. A
shared encoder $\enc$ embeds two such crops and a small head scores the pair,
$\judge(a,\refc)$. For output frame $f$, the reference $\refc_{j(f)}$ is one of the
sixteen references the generator received in the same step. With $y$ the real frames,
$\tilde x$ the composited prediction, $y'_f$ a real frame of another person, and
$\mathrm{sp}(u)=\log(1+e^{u})$, the judge minimizes
\begin{equation}
\mathcal{L}_{J}(\psi)=\frac1F\sum_{f=1}^{F}\Big[
\underbrace{\mathrm{sp}\big(-\judge(y_f,\refc_{j(f)})\big)}_{\text{same person, real}}
+\underbrace{\mathrm{sp}\big(\judge(\tilde x_f,\refc_{j(f)})\big)}_{\text{generated}}
+\underbrace{\mathrm{sp}\big(\judge(y'_{f},\refc_{j(f)})\big)}_{\text{other person, real}}
\Big],
\label{eq:judge_loss}
\end{equation}
while the generator minimizes the non-saturating pair term
\begin{equation}
\mathcal{L}_{\mathrm{pair}}(\theta)=\frac1F\sum_{f=1}^{F}
\mathrm{sp}\big(-\judge(\tilde x_f,\refc_{j(f)})\big),
\label{eq:gen_loss}
\end{equation}
with $\psi$ frozen, back-propagated through $\crop$, the composite of
Eq.~\ref{eq:composite}, and $\mathcal{D}$. Because a real mouth of another person is a
negative, a realistic but generic mouth does not lower the generator's cost; only a
mouth that the judge cannot tell apart from the referenced person's real mouth does. The
judge is trained together with the generator.

\subsection{Shape term}\label{sec:shape}
The shape term supervises how far the mouth opens and how much of the lips shows. A
frozen face parser $\parser$ \citep{p_segformer,p_celebamaskhq} runs on the composite
with gradient, and its soft mouth-interior and lip probabilities are compared with the
parser's confident regions on ground truth. With $p^{\mathrm{mouth}}$ and
$p^{\mathrm{lip}}$ the parser's soft mouth-interior and lip probabilities on $\tilde x$
(the lip probability covering both lips), $M^{\mathrm{mouth}}$ and $M^{\mathrm{lip}}$
the corresponding confident supports of $\parser(y)$, $\varrho=\mathbf{1}[\alpha>0]$
the editable region of Eq.~\ref{eq:composite}, and $\lambda$ a fixed weight,
\begin{equation}
\begin{aligned}
\mathrm{Dice}(q,M)&=1-\frac{2\sum\varrho\,q\,M+1}{\sum\varrho\,q+\sum\varrho\,M+1},\\
\mathcal{L}_{\mathrm{shape}}&=\mathrm{Dice}(p^{\mathrm{mouth}},M^{\mathrm{mouth}})
+\lambda\,\mathrm{Dice}(p^{\mathrm{lip}},M^{\mathrm{lip}}).
\end{aligned}
\label{eq:dice}
\end{equation}
A Dice score does not reward blurring, so this term constrains the geometry of the mouth
without pulling its texture toward an average.

\subsection{Data}\label{sec:data}
We build our data from the frontal camera of NeRSemble
\citep{kirschstein2023nersemble}, a multi-view recording in which each person reads ten
sentences and performs lip, mouth, tongue, and jaw expression sequences. We split 340
of its identities by person into a 290-identity training corpus and a disjoint
50-identity held-out benchmark (Sec.~\ref{sec:protocol}). All clips are aligned to a
$512{\times}512$ canvas, and the corpus provides 2{,}900 sentence clips, cut into about
18.6k units of 16 frames.
Each identity contributes two parts:
its sentence clips, which are the videos to be re-rendered, and an enrollment bank of
other clips of the same person, which never serve as queries and supply the
references (Fig.~\ref{fig:data}). A
training sample is built from one identity and consists of five aligned parts: a
16-frame window of one of its clips with the lower face masked, the audio of that
window, one reference frame per output frame drawn from the same identity's bank, the
four HD patches of that identity, and the unmasked window as the target.

\subsection{Training and inference}\label{sec:objective}
The training objective adds the pair term of Eq.~\ref{eq:gen_loss} and the shape term
of Eq.~\ref{eq:dice} to three image terms: LPIPS on the composited frame, a boundary
term, and DISTS on the oral box. We fine-tune the parent with this
objective. At inference, each clip
is denoised in 20 steps as one latent sequence: at every step the model runs on
overlapping 16-frame windows whose predictions are fused
\citep{p_multidiffusion,p_genlvideo}, so consecutive windows join without seams.

\section{Evaluation Protocol}\label{sec:protocol}
\begin{table}[t]
\centering\scriptsize
\setlength{\tabcolsep}{0pt}
\renewcommand{\arraystretch}{0.88}
\caption{Held-out NeRSemble benchmark. \dag{}: run as released, with the target frame as reference.}
\label{tab:main}
\begin{tabular*}{\linewidth}{@{\extracolsep{\fill}}lcccccccc@{}}
\toprule
Method (condition) & MAE$\downarrow$ & MAE-out$\downarrow$ & PSNR$\uparrow$ & LPIPS$\downarrow$ & DISTS$\downarrow$ & Tex$\to$1 & Open$\to$1 & LSE-C$\uparrow$/D$\downarrow$ \\
\midrule
Wav2Lip (xref) & 0.0463 & 0.0285 & 23.7 & 0.473 & 0.340 & 0.200 & 0.493 & 4.28/8.19 \\
LatentSync-1.6 (xref) & \textbf{0.0328} & 0.0183 & \textbf{26.1} & 0.334 & 0.234 & 0.399 & 0.451 & 5.12/6.54 \\
MuseTalk-1.5 (official)$^\dag$ & 0.0440 & 0.0150 & 23.7 & 0.446 & 0.324 & 0.338 & 0.628 & 3.20/8.69 \\
X-Dub (xvid) & 0.0483 & 0.0460 & 22.5 & 0.408 & 0.251 & 0.555 & 1.42 & 3.90/9.11 \\
sync.so lipsync-2 (xvid) & 0.0466 & 0.0447 & 23.1 & 0.366 & 0.207 & 0.698 & 0.860 & 1.70/11.1 \\
Kling lip-sync (xvid) & 0.0502 & 0.0475 & 22.4 & 0.354 & 0.212 & 0.704 & \textbf{1.09} & 3.13/9.33 \\
HeyGen lip-sync (xvid) & 0.0423 & 0.0430 & 24.0 & 0.373 & 0.227 & 0.437 & 0.653 & 5.37/7.13 \\
\midrule
\ours{} (ours) & 0.0342 & \textbf{0.0117} & 25.5 & \textbf{0.297} & \textbf{0.163} & \textbf{0.833} & \textbf{0.880} & 4.24/6.56 \\
\midrule
Ground truth & -- & -- & -- & -- & -- & -- & -- & 4.02/8.26 \\
\bottomrule
\end{tabular*}
\end{table}
The held-out NeRSemble benchmark holds 50 identities never used in training; each has
two query sentences, which a system must reproduce (100 query clips), and an enrollment
bank of two other sentences, which it may use. Every system is scored against the target
clip, and apart from MuseTalk-1.5, which we run as released, none is shown that clip's
mouth. The mouth-inpainting systems (Wav2Lip, LatentSync-1.6, and \ours{}) receive the
target frames with the lower face masked and reference frames from other clips of the
same person (\textbf{xref}). The cross-video systems (X-Dub \citep{he2025xdub} and the commercial services)
cannot take a masked input; they re-render another clip of the person with the target
audio (\textbf{xvid}). The clips of a person are recorded in one session with the same frontal
camera, so this clip is close to the target in head position and expression. The released inference code of Wav2Lip, LatentSync, and MuseTalk uses the
unmasked copy of the frame being edited as the reference, which in a paired test shows
the model the answer (Appendix~\ref{app:leak}); for the xref rows we replace only this
reference input. Fidelity is measured on the fixed oral box with MAE, PSNR, LPIPS
\citep{zhang2018lpips}, DISTS \citep{ding2021dists}, a texture ratio and an open-area ratio
(the inner mouth's high-frequency energy and the parser's mouth-interior area, each
divided by ground truth's, both best at one), synchronization with LSE-C/LSE-D from the
official SyncNet \citep{chung2016syncnet,p_wav2lip}; MAE-out, the error over the rest
of the frame, measures how much a system changes outside the mouth
(Appendix~\ref{app:protocol}).

\section{Experiments}\label{sec:experiments}

\subsection{Experimental Setup}\label{sec:setup}
We evaluate all systems on the held-out NeRSemble benchmark under the protocol of
Sec.~\ref{sec:protocol}, and compare with Wav2Lip \citep{p_wav2lip}, LatentSync-1.6 \citep{li2024latentsync},
MuseTalk-1.5 \citep{zhang2025musetalk}, the public X-Dub release \citep{he2025xdub}, a
mask-free model built on Wan2.2-TI2V-5B, and the commercial services sync.so lipsync-2, Kling
lip-sync, and HeyGen lip-sync \citep{p_syncso,p_kling,p_heygen}. Each
baseline is run from its released weights or public API with its officially
recommended inference setting; the protocol changes only the inputs it is given
(Appendix~\ref{app:protocol}).

\subsection{Results}\label{sec:main}
\ifdefined\qpx\else\newlength{\qpx}\fi
\begin{figure}[t]
\centering
\setlength{\qpx}{\dimexpr\linewidth/1768\relax}%
\begin{tikzpicture}[x=\qpx,y=\qpx,
  qh/.style={font=\sffamily\tiny,anchor=south,align=center,inner sep=0pt}]
\node[anchor=north west,inner sep=0pt] at (0,0)
  {\animategraphics[loop,width=\linewidth]{2}{figures/qualitative/anim/q_}{0}{7}};
\node[qh] at (96,12)   {\textbf{LatentSync}\\1.6 (xref)\strut};
\node[qh] at (292,12)  {\textbf{MuseTalk}\\1.5 (official)\strut};
\node[qh] at (488,12)  {\textbf{Wav2Lip}\\(xref)\strut};
\node[qh] at (684,12)  {\textbf{X-Dub}\\(xvid)\strut};
\node[qh] at (880,12)  {\textbf{sync.so}\\(xvid)\strut};
\node[qh] at (1076,12) {\textbf{Kling}\\(xvid)\strut};
\node[qh] at (1272,12) {\textbf{HeyGen}\\(xvid)\strut};
\node[qh] at (1476,12) {\textbf{\ours{}}\\(ours, xref)\strut};
\node[qh] at (1672,12) {\textbf{Ground truth}\\\strut};
\end{tikzpicture}
\caption{Qualitative comparison; click to play the video clips in \textcolor{red}{Adobe Acrobat}.}
\label{fig:qual}
\end{figure}

Fig.~\ref{fig:teaser} and Fig.~\ref{fig:qual} show \ours{} on held-out identities. A key
strength of our approach is that the rendered mouth is recognizably the person's own.
From a video whose lower face is masked and a few enrollment frames from other
recordings of the same person, \ours{} reconstructs the contour and texture of the lips and the shape and
arrangement of the individual teeth, instead of falling back to a generic dentition.
The teeth are rendered with crisp, well-separated boundaries, and the lips keep their
natural texture rather than the smoothed appearance typical of inpainting-based
dubbing. The mouth opens and closes in step with the speech, and since only the masked
lower face is regenerated and composited back, everything outside it is kept from the
source video, including the eyes, skin texture, hair, and expression. These
properties are consistent across the benchmark: the oral DISTS of \ours{} is lower than
that of every baseline on 95 of the 100 query clips.

\subsection{Comparison and Evaluation}\label{sec:compare}
\paragraph{Quantitative evaluation.}
Table~\ref{tab:main} reports the held-out benchmark, and the radar of
Fig.~\ref{fig:teaser} summarizes it. \ours{} achieves the best oral LPIPS and DISTS of
all systems, open-source and commercial, so the mouth it renders is the closest to the
real one in perceptual structure and texture; its DISTS is lower even than that of
LatentSync-1.6 given the target frame itself as its reference (Appendix~\ref{app:leak}).
Its texture ratio is the closest to one: the high-frequency detail of the inner mouth,
which carries the tooth boundaries and gaps, is preserved rather than smoothed away.
Among the masked inpainters, its open-area ratio is also the closest to one, so the
mouth opens about as widely as the person's does, and it has the lowest error outside
the oral box, editing only the region it must. On pixel-averaging metrics \ours{} is a
close second to LatentSync-1.6. Such metrics favour a smooth, averaged mouth, as the low
texture ratio of LatentSync-1.6 shows, and a sharper, more detailed mouth pays a small
pixel-level price for its fidelity. The cross-video systems re-render a complete,
unmasked recording of the person, so every frame they output starts from the person's
real mouth; even so, \ours{} surpasses all of them on every perceptual metric. For
synchronization, \ours{} scores better than ground truth's own recordings on both
LSE-C and LSE-D, placing it on the synchronized side of real video.

\begin{table}[t]
\centering\scriptsize
\setlength{\tabcolsep}{0pt}
\renewcommand{\arraystretch}{0.88}
\caption{Ablation on the held-out benchmark (scored on lossless renders).}
\label{tab:ablation}
\begin{tabular*}{\linewidth}{@{\extracolsep{\fill}}lcccccccc@{}}
\toprule
Variant & MAE$\downarrow$ & MAE-out$\downarrow$ & PSNR$\uparrow$ & LPIPS$\downarrow$ & DISTS$\downarrow$ & Tex$\to$1 & Open$\to$1 & LSE-C$\uparrow$/D$\downarrow$ \\
\midrule
w/o scattered references & 0.0407 & 0.00992 & 24.7 & 0.311 & 0.184 & \textbf{0.869} & 0.777 & 4.28/6.49 \\
w/o HD patches & 0.0364 & 0.00639 & 25.2 & 0.295 & 0.165 & 0.815 & 0.740 & 4.06/6.52 \\
w/o paired judge & 0.0367 & 0.00763 & 25.3 & 0.326 & 0.185 & 0.778 & 0.430 & 3.86/6.95 \\
w/o shape term & 0.0395 & 0.00781 & 24.7 & 0.317 & 0.181 & 0.842 & 0.702 & 3.77/6.79 \\
\midrule
\ours{} (full) & \textbf{0.0352} & \textbf{0.00634} & \textbf{25.4} & \textbf{0.291} & \textbf{0.161} & 0.858 & \textbf{0.858} & 4.24/6.57 \\
\bottomrule
\end{tabular*}
\end{table}
\ifdefined\abpx\else\newlength{\abpx}\fi
\expandafter\def\csname abannot@0\endcsname{\abtitle{388}{4}{w/o scattered references}%
  \absub{128}{61}{Variant}%
  \absub{388}{61}{Ours}%
  \absub{648}{61}{GT}%
  \abbox{abLipSync}{73}{203}{179}{255}%
  \ablabel{abLipSync}{73}{255}{north west}{lip-sync error}%
  \abtitle{1190}{4}{w/o HD patches}%
  \absub{930}{61}{Variant}%
  \absub{1190}{61}{Ours}%
  \absub{1450}{61}{GT}%
  \abbox{abDetail}{899}{202}{960}{223}%
  \ablabel{abDetail}{899}{202}{south west}{detail loss}%
  \abtitle{388}{342}{w/o paired judge}%
  \absub{128}{399}{Variant}%
  \absub{388}{399}{Ours}%
  \absub{648}{399}{GT}%
  \abbox{abIdentityDrift}{75}{536}{183}{582}%
  \ablabel{abIdentityDrift}{75}{536}{south west}{identity drift}%
  \abtitle{1190}{342}{w/o shape term}%
  \absub{930}{399}{Variant}%
  \absub{1190}{399}{Ours}%
  \absub{1450}{399}{GT}%
  \abbox{abAmplitude}{884}{543}{978}{590}%
  \ablabel{abAmplitude}{884}{590}{north west}{opening amplitude}}
\expandafter\def\csname abannot@1\endcsname{\abtitle{388}{4}{w/o scattered references}%
  \absub{128}{61}{Variant}%
  \absub{388}{61}{Ours}%
  \absub{648}{61}{GT}%
  \abbox{abLipSync}{73}{203}{179}{255}%
  \ablabel{abLipSync}{73}{255}{north west}{lip-sync error}%
  \abtitle{1190}{4}{w/o HD patches}%
  \absub{930}{61}{Variant}%
  \absub{1190}{61}{Ours}%
  \absub{1450}{61}{GT}%
  \abbox{abDetail}{899}{202}{960}{223}%
  \ablabel{abDetail}{899}{202}{south west}{detail loss}%
  \abtitle{388}{342}{w/o paired judge}%
  \absub{128}{399}{Variant}%
  \absub{388}{399}{Ours}%
  \absub{648}{399}{GT}%
  \abbox{abIdentityDrift}{75}{536}{183}{582}%
  \ablabel{abIdentityDrift}{75}{536}{south west}{identity drift}%
  \abtitle{1190}{342}{w/o shape term}%
  \absub{930}{399}{Variant}%
  \absub{1190}{399}{Ours}%
  \absub{1450}{399}{GT}%
  \abbox{abAmplitude}{884}{543}{978}{590}%
  \ablabel{abAmplitude}{884}{590}{north west}{opening amplitude}}
\expandafter\def\csname abannot@2\endcsname{\abtitle{388}{4}{w/o scattered references}%
  \absub{128}{61}{Variant}%
  \absub{388}{61}{Ours}%
  \absub{648}{61}{GT}%
  \abbox{abLipSync}{73}{203}{179}{255}%
  \ablabel{abLipSync}{73}{255}{north west}{lip-sync error}%
  \abtitle{1190}{4}{w/o HD patches}%
  \absub{930}{61}{Variant}%
  \absub{1190}{61}{Ours}%
  \absub{1450}{61}{GT}%
  \abinset{842}{242}{1018}{318}%
  \abinset{1102}{242}{1278}{318}%
  \abinset{1362}{242}{1538}{318}%
  \abbox{abDetail}{899}{202}{960}{223}%
  \ablabel{abDetail}{899}{202}{south west}{detail loss}%
  \abtitle{388}{342}{w/o paired judge}%
  \absub{128}{399}{Variant}%
  \absub{388}{399}{Ours}%
  \absub{648}{399}{GT}%
  \abinset{47}{586}{209}{656}%
  \abinset{307}{586}{469}{656}%
  \abinset{567}{586}{729}{656}%
  \abbox{abIdentityDrift}{75}{536}{183}{582}%
  \ablabel{abIdentityDrift}{75}{536}{south west}{identity drift}%
  \abtitle{1190}{342}{w/o shape term}%
  \absub{930}{399}{Variant}%
  \absub{1190}{399}{Ours}%
  \absub{1450}{399}{GT}%
  \abbox{abAmplitude}{884}{543}{978}{590}%
  \ablabel{abAmplitude}{884}{590}{north west}{opening amplitude}}
\expandafter\def\csname abannot@3\endcsname{\abtitle{388}{4}{w/o scattered references}%
  \absub{128}{61}{Variant}%
  \absub{388}{61}{Ours}%
  \absub{648}{61}{GT}%
  \abbox{abLipSync}{73}{203}{179}{255}%
  \ablabel{abLipSync}{73}{255}{north west}{lip-sync error}%
  \abtitle{1190}{4}{w/o HD patches}%
  \absub{930}{61}{Variant}%
  \absub{1190}{61}{Ours}%
  \absub{1450}{61}{GT}%
  \abbox{abDetail}{899}{202}{960}{223}%
  \ablabel{abDetail}{899}{202}{south west}{detail loss}%
  \abtitle{388}{342}{w/o paired judge}%
  \absub{128}{399}{Variant}%
  \absub{388}{399}{Ours}%
  \absub{648}{399}{GT}%
  \abbox{abIdentityDrift}{75}{536}{183}{582}%
  \ablabel{abIdentityDrift}{75}{536}{south west}{identity drift}%
  \abtitle{1190}{342}{w/o shape term}%
  \absub{930}{399}{Variant}%
  \absub{1190}{399}{Ours}%
  \absub{1450}{399}{GT}%
  \abbox{abAmplitude}{884}{543}{978}{590}%
  \ablabel{abAmplitude}{884}{590}{north west}{opening amplitude}}
\def\abannot#1{\csname abannot@#1\endcsname}

\begin{figure}[t]
\centering
\setlength{\abpx}{\dimexpr\linewidth/1578\relax}%
\definecolor{abLipSync}{RGB}{225,20,20}%
\definecolor{abIdentityDrift}{RGB}{255,140,0}%
\definecolor{abAmplitude}{RGB}{220,0,200}%
\definecolor{abDetail}{RGB}{0,170,220}%
\newcommand{\abbox}[5]{\draw[#1,line width=0.5pt] (#2,#3) rectangle (#4,#5);}%
\newcommand{\abinset}[4]{\draw[white,line width=0.6pt] (#1,#2) rectangle (#3,#4);}%
\expandafter\def\csname abtextabLipSync\endcsname{white}\expandafter\def\csname abtextabIdentityDrift\endcsname{black}%
\expandafter\def\csname abtextabAmplitude\endcsname{white}%
\expandafter\def\csname abtextabDetail\endcsname{black}%
\newcommand{\ablabel}[5]{\node[anchor=#4,inner sep=0.6pt,outer sep=0pt,fill=#1,text=\csname abtext#1\endcsname,
  font=\sffamily\fontsize{4.5}{5}\selectfont] at (#2,#3) {#5};}%
\newcommand{\abtitle}[3]{\node[anchor=north,inner sep=0pt,font=\sffamily\scriptsize\bfseries] at (#1,#2) {#3};}%
\newcommand{\absub}[3]{\node[anchor=base,inner sep=0pt,font=\sffamily\tiny] at (#1,#2) {#3};}%
\newcommand{\abframe}[1]{\begin{tikzpicture}[x={(\abpx,0)},y={(0,-\abpx)}]%
  \useasboundingbox (0,0) rectangle (1578,658);%
  \node[anchor=north west,inner sep=0pt,outer sep=0pt] at (0,0) {\includegraphics[width=\linewidth]{figures/ablation/abl_0#1.jpg}};%
  \abannot{#1}%
  \end{tikzpicture}}%
\noindent\begin{animateinline}[loop,poster=2]{2}
\abframe{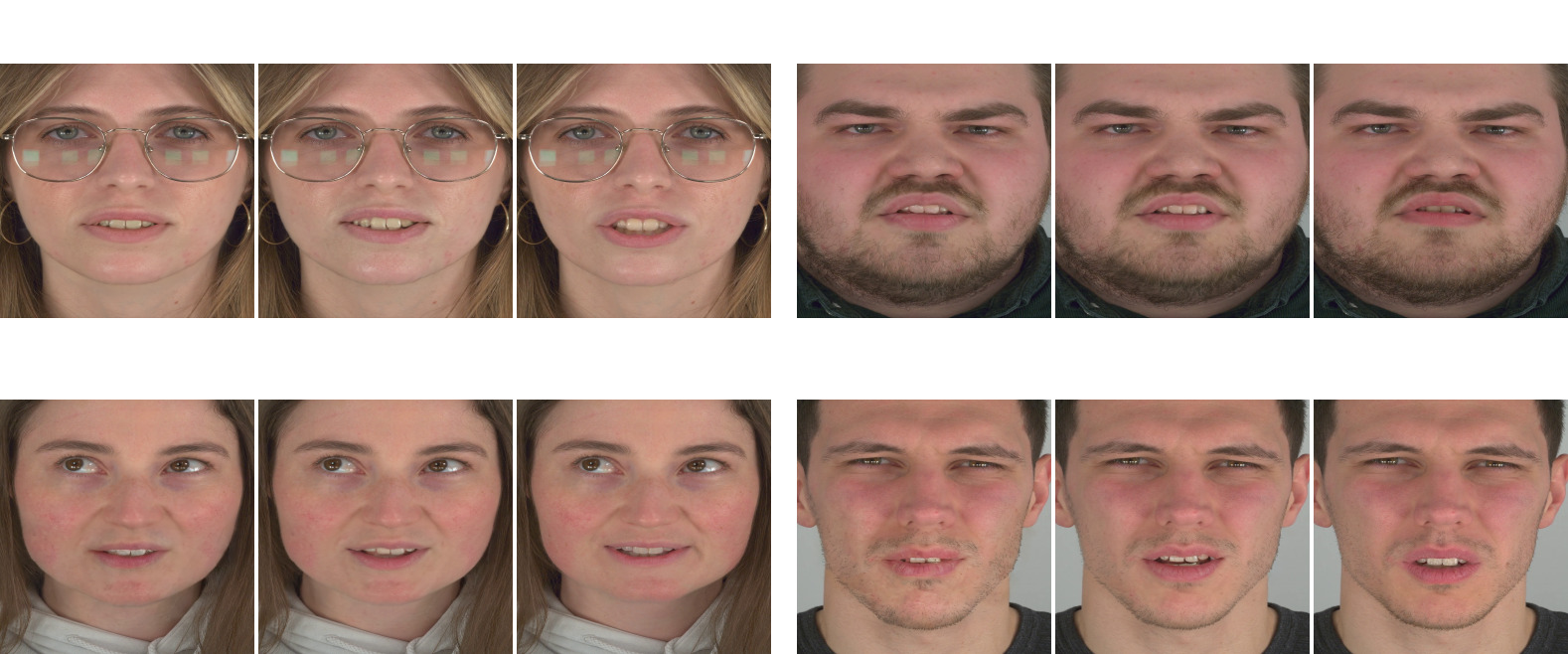}\newframe\abframe{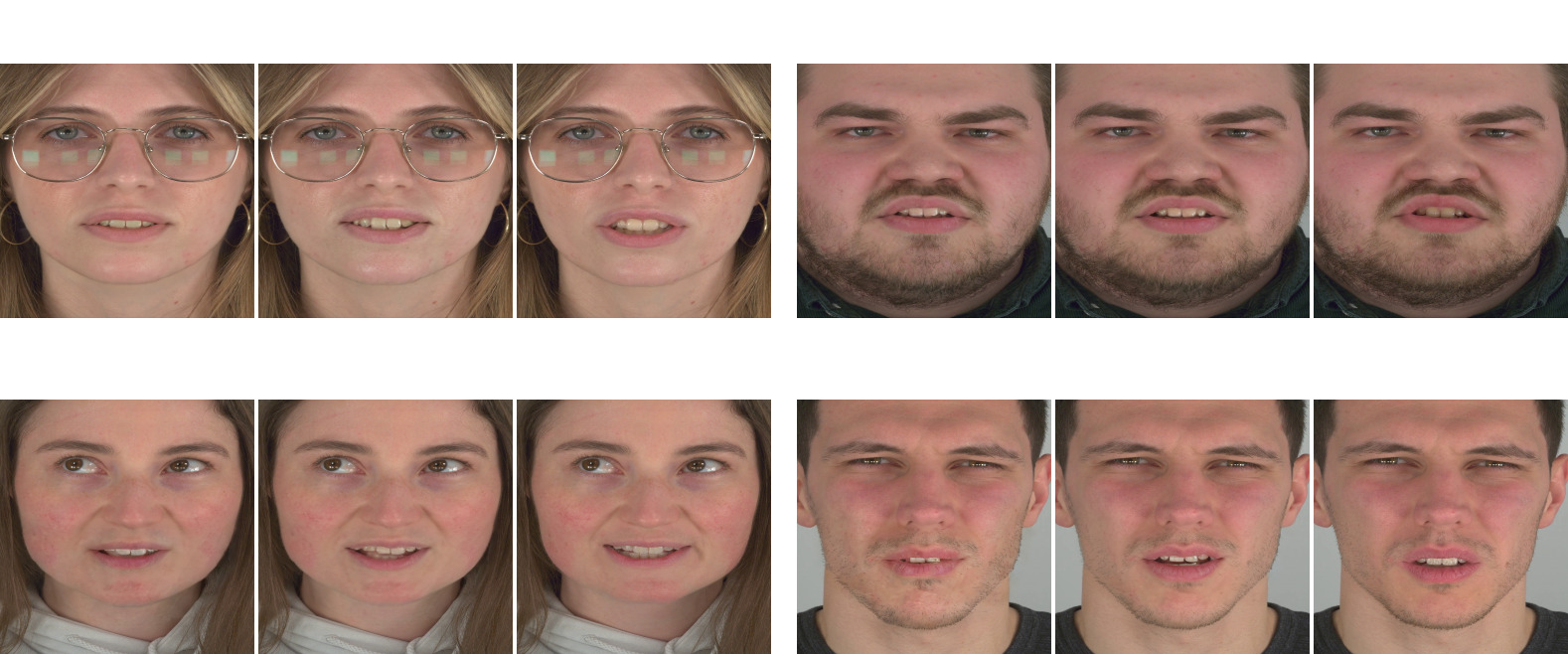}\newframe\abframe{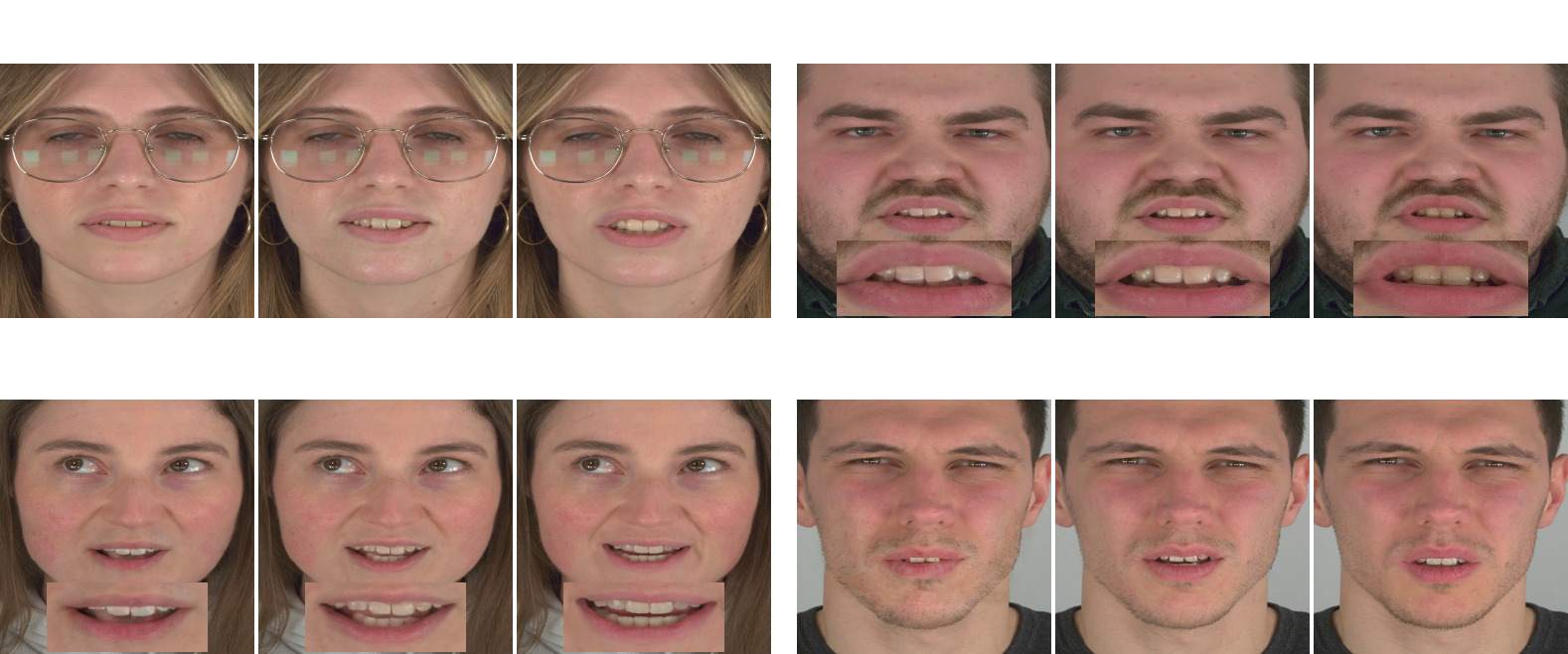}\newframe\abframe{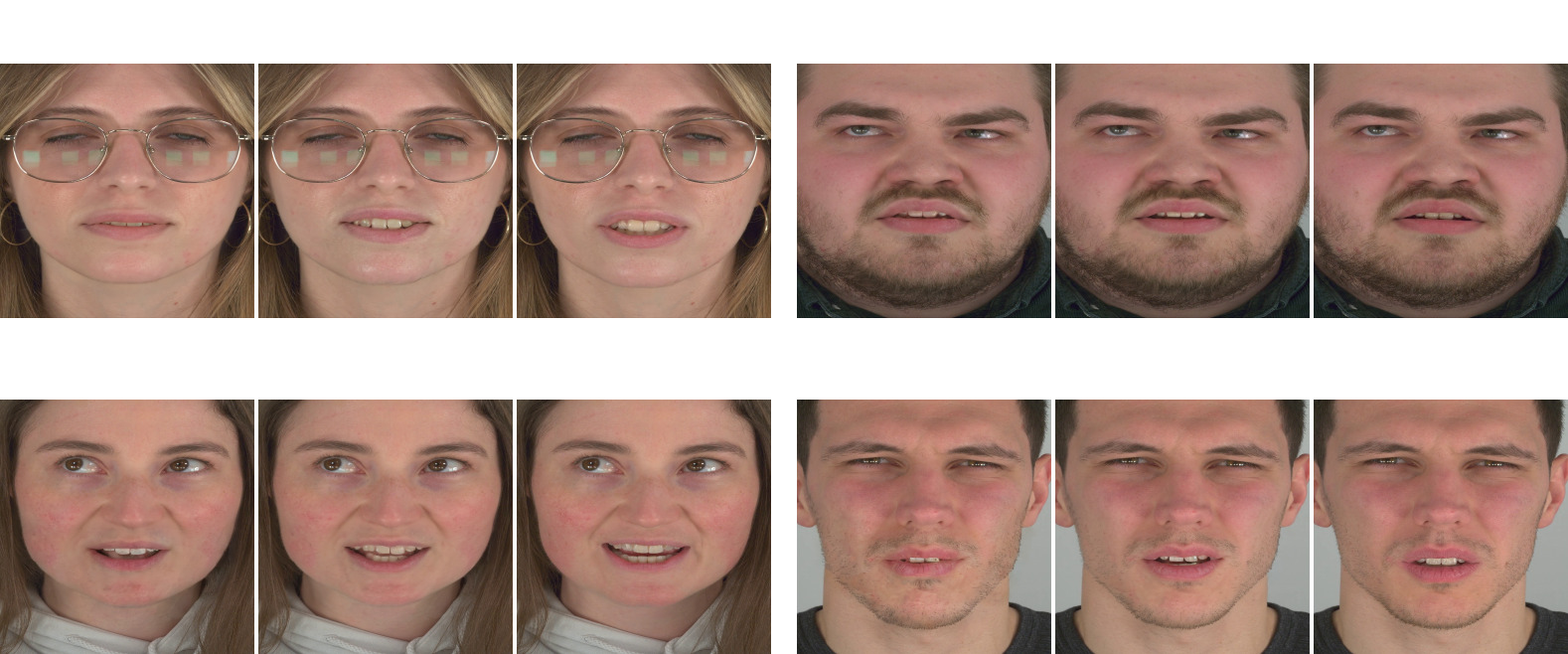}
\end{animateinline}
\caption{Ablation study; boxes mark the failures; click to play the video clips in \textcolor{red}{Adobe Acrobat}.}
\label{fig:ablation}
\end{figure}

\paragraph{Qualitative evaluation.}
Fig.~\ref{fig:qual} compares all systems on the same frame of four held-out identities.
The inpainting baselines lose the person's dentition: Wav2Lip and MuseTalk-1.5 produce
blurred mouths in which individual teeth are hard to discern, and LatentSync-1.6
renders a smooth, generic row of teeth. X-Dub tends to open the mouth too wide, and the
commercial services show lips and teeth that differ from the ground truth at the same frame. In
contrast, \ours{} produces sharp, well-separated teeth whose shape and arrangement follow
the person's enrollment, lips with a natural contour and texture, and an opening that
matches the ground truth, blended seamlessly into the untouched face. The difference is
clearest in the mouth close-ups, where the teeth of \ours{} most closely resemble the
person's real ones.

\subsection{Ablation Study}\label{sec:ablation}
We ablate \ours{} on the held-out benchmark by removing one component at a time
(Table~\ref{tab:ablation} and Fig.~\ref{fig:ablation}). To compare fine detail,
Table~\ref{tab:ablation} is scored on lossless renders, so the full model's values differ
slightly from Table~\ref{tab:main}. The full model is the best or
close to the best on every metric, and each removal damages the mouth in its own way,
which shows that the components are not a stack of independent additions but parts of
one mechanism, each of which the others rely on.

\paragraph{Without scattered references.} This variant conditions every frame on a
contiguous window of enrollment frames instead of frames scattered over the whole bank.
Such a window covers only the few mouth shapes of one short stretch of speech, so the
generator copies the opening of that stretch rather than following the audio: in
Fig.~\ref{fig:ablation} its mouth closes while the person keeps it open, and in
Table~\ref{tab:ablation} its mouth is less faithful and opens less. Scattered frames
show the person's mouth in many opening states, from which the generator takes its
appearance for any opening the audio calls for.

\paragraph{Without the HD patches.} This variant removes the HD patches from training and
inference, so the person's mouth reaches the generator only through references encoded
by the VAE. Fine detail is the first to go: in Fig.~\ref{fig:ablation} the variant opens
the mouth to a similar extent but renders the teeth as smooth, uniform shapes with soft
outlines, and even more detail is lost on the lips and the surrounding skin, where the
creases of the lower lip and the stubble flatten. In Table~\ref{tab:ablation} the texture
ratio drops further below one, the perceptual scores fall behind, and the mouth opens
less. The HD patches are the only path by which the person's teeth, lips, and skin reach
the generator at the resolution of the video.

\paragraph{Without the paired judge.} The judge compares the rendered mouth with the
person's reference and rejects a realistic mouth of someone else, so it is the part of
the objective that asks the mouth to be this person's. Without it, the image terms are
satisfied by any plausible mouth, and the teeth drift toward a generic dentition: in
Fig.~\ref{fig:ablation} the variant renders a flat, uniform row of upper teeth where the
person shows both tooth rows with their own shapes, and in Table~\ref{tab:ablation} its
perceptual scores and texture ratio fall behind the full model.

\paragraph{Without the shape term.} The shape term matches the mouth interior and the
lips of the rendered frame to those of the real frame, so it is the part of the
objective that fixes how far the mouth opens and how the lips are shaped. Without it,
the opening is left to the audio prior, which under-articulates: in
Fig.~\ref{fig:ablation} the variant opens the mouth at the right moments but only about
half as wide as the person does, and its open-area ratio in Table~\ref{tab:ablation}
falls well below one.

\subsection{Human Study}\label{sec:human}
\begin{figure}[t]
\centering
\includegraphics{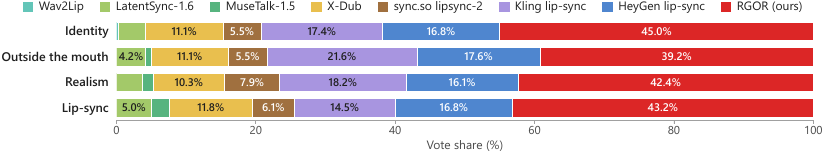}%
\caption{Human study: share of votes each system receives as the best result on each question.}
\label{fig:userstudy}
\end{figure}

We conduct a human study to evaluate the rendered mouths from a human perspective. Each
screen shows the person's real recording as the reference and the eight systems below
it, anonymized and shuffled, and nineteen raters pick the best video on four criteria:
mouth identity, the face outside the mouth, realism, and lip sync
(Appendix~\ref{app:human}). As shown in Fig.~\ref{fig:userstudy}, \ours{} is clearly
preferred on all four. The strongest preference is on mouth identity, where raters
recognize the person's own teeth and lip shape, which the generic mouths of the other
systems lack. The preference on realism
and on the face outside the mouth follows from a sharp, detailed mouth blended into an
untouched face, and the equally clear preference on lip sync shows that grounding the
mouth in the person's references does not come at the cost of following the speech.
These judgments agree with the perceptual metrics.

\section{Conclusion}
We presented \ours{}, which makes an audio-driven lip-sync generator render the specific
person's mouth by grounding it in that person's enrollment references and HD patches and training it against
a paired judge. Under a protocol that withholds the
target mouth from every system that accepts a separate reference, \ours{} ranks first or second on
every fidelity metric of Table~\ref{tab:main} against open-source and commercial systems, raters
prefer it on every question of the human study, and it preserves the person's own lip and dental
detail while leaving the rest of the face untouched. These results come from four
components working as one: the scattered references show the person's mouth in many shapes
without a motion to copy, the HD patches carry its fine detail past the VAE, the paired
judge steers the generator toward this person's mouth, and the shape term keeps the opening
in step with the real frame; removing any one of them degrades the mouth in its own way.
We believe that the principle behind it, a
reference that the objective must account for and a judge that compares against it, can extend
to other person-specific fine structure, such as hands, eyes, and hair. Failure cases
and limitations are analyzed in Appendix~\ref{app:failure}.

\bibliography{references}
\bibliographystyle{iclr2027_conference}
\clearpage
\appendix
\section{Human Study}\label{app:human}
\begin{figure}[p]
\centering
\includegraphics[width=\linewidth]{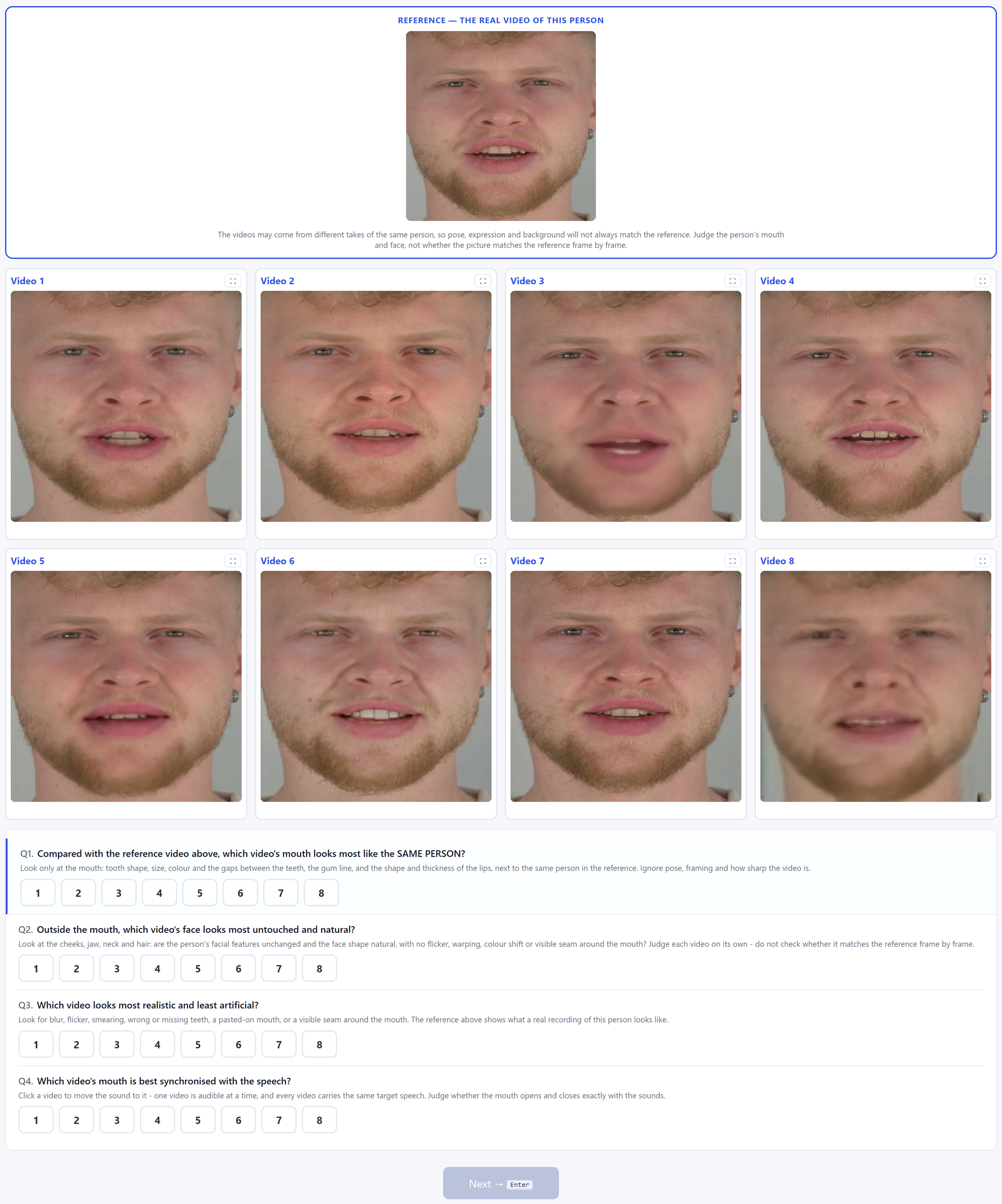}
\caption{The human-study interface for one held-out query clip.}
\label{fig:ui}
\end{figure}

\paragraph{Interface.} Fig.~\ref{fig:ui} shows one screen of the study. The person's
real recording of the sentence plays at the top as the reference, and the eight systems
play below it in a grid labelled only Video~1 to Video~8, in an order reshuffled for
every rater and screen. All videos loop muted, and the sound moves to whichever video
the rater selects, so every candidate is heard with the same target speech. Raters are
told that a candidate may come from a different take of the same person, so that pose
and background can differ, and that the mouth and the face are what is judged.

\paragraph{Questions.} For each screen, raters pick one video for each of four
questions:
\begin{itemize}
\item Identity: ``Compared with the reference video above, which video's mouth looks
most like the SAME PERSON?''
\item Outside: ``Outside the mouth, which video's face looks most untouched and
natural?''
\item Realism: ``Which video looks most realistic and least artificial?''
\item Lip-sync: ``Which video's mouth is best synchronised with the speech?''
\end{itemize}

\paragraph{Participants.} The study covers 40 held-out identities, one query clip each,
and each rater judges a random 20 of them. Twenty-two raters without prior exposure to
the project took part. We removed three raters whose answers were not usable: one
chose the same video for all four questions on almost every screen, and two answered
in a few seconds per screen, faster than the videos can be watched. The remaining 19
raters give 380 votes per question.

\paragraph{Agreement with the automatic metrics.} To verify that the metrics of our
evaluation protocol reflect human judgment, Table~\ref{tab:corr} gives, for each
question and metric, the Spearman rank correlation between the vote shares of the eight
systems and their metric values, oriented so that a positive value means that the
metric ranks the systems as the raters do. The perceptual and texture metrics on the
oral box correlate positively with the raters on every question, DISTS and Tex
significantly, whereas the pixel-averaging metrics do
not: LatentSync-1.6, which leads on MAE and PSNR, receives few votes, because raters
prefer a sharp, detailed mouth to an averaged one.

\begin{table}[h]
\centering\footnotesize
\setlength{\tabcolsep}{5pt}
\caption{Spearman correlation between the human vote shares of the eight systems and the
metrics of Table~\ref{tab:main}; $^*$: $p<0.05$ (exact permutation test). The eight
systems receive the same vote ranking on Outside the mouth and on Realism, so these two
columns coincide.}
\label{tab:corr}
\begin{tabular}{@{}lcccc@{}}
\toprule
Metric & Identity & Outside the mouth & Realism & Lip-sync \\
\midrule
MAE & $-$0.0952 & $-$0.0714 & $-$0.0714 & 0.0476 \\
MAE-out & $-$0.238 & $-$0.190 & $-$0.190 & $-$0.119 \\
PSNR & $-$0.0714 & $-$0.0952 & $-$0.0952 & 0.0238 \\
LPIPS & 0.667 & 0.690 & 0.690 & 0.643 \\
DISTS & 0.786$^*$ & 0.810$^*$ & 0.810$^*$ & 0.786$^*$ \\
Tex$\to$1 & 0.881$^*$ & 0.905$^*$ & 0.905$^*$ & 0.833$^*$ \\
Open$\to$1 & 0.714 & 0.762$^*$ & 0.762$^*$ & 0.690 \\
\bottomrule
\end{tabular}
\end{table}

\section{Protocol and Metrics}\label{app:protocol}
\paragraph{Benchmark.} The 50 held-out NeRSemble identities are never used for training.
Each contributes two query sentences (100 query clips) and an enrollment bank of two
further sentences; a bank clip is never a query, so an enrollment bank never contains
the target clip's mouth. HDTF (Appendix~\ref{app:hdtf}) has 24 in-the-wild
speakers with two clips each; each clip is a query once, with the other clip as its
enrollment (48 query clips).

\paragraph{Conditions.} The mouth inpainters (Wav2Lip, LatentSync-1.6, and \ours{}) run
in xref: they receive the target clip with the lower face masked, the target audio, and
reference frames from the person's enrollment bank (for \ours{}, also the four HD patches), so their output is
frame-aligned with the target. X-Dub and the commercial services cannot take a masked
input and run in xvid: they re-render an enrollment clip of the same person with the
target audio. This clip is recorded in the same session with the same frontal camera, so
its head position and expression are close to those of the target. MuseTalk-1.5
is run as released, with the target frame as its reference (\dag{}). For the xref rows of Wav2Lip and LatentSync-1.6 we replace only
the reference input of the released code, and every baseline otherwise uses its
officially recommended inference setting.

\paragraph{Metrics.} Each output is aligned in time with the target clip and compared
with it frame by frame in two regions: the oral box, a fixed rectangle around the mouth of the aligned canvas that is
the same for every identity, system, and frame, and the rest of the frame. Inside the
oral box we compute MAE, PSNR, LPIPS \citep{zhang2018lpips}, DISTS
\citep{ding2021dists}, and the two ratios below; outside it we compute the same pixel
error, MAE-out. LSE-C and LSE-D use the public SyncNet evaluator
\citep{chung2016syncnet,p_wav2lip} with the settings of the official LatentSync
evaluation.

\paragraph{The two ratios.} Both ratios compare the output with ground truth inside the
oral box and are best at one. The texture ratio (Tex) divides the high-frequency energy
of the output's inner mouth by that of ground truth: below one for a smoothed or blurred
mouth, above one for an over-textured one. The open-area ratio (Open) divides the
mouth-interior area that a face parser finds in the output by that in ground truth:
below one for a mouth held too closed, above one for one opened too wide.

\section{Same-Frame Reference Leak}\label{app:leak}
The released inference code of Wav2Lip, LatentSync, and MuseTalk takes a single video
and uses the unmasked copy of each frame it edits as that frame's reference, so in paired
reconstruction the model is shown the answer. Table~\ref{tab:leak} runs Wav2Lip and
LatentSync-1.6 both ways: with a reference from a different clip of the same person
(xref, the rows of Table~\ref{tab:main}) and as released (official). With the answer
as its reference, LatentSync-1.6's oral LPIPS drops from 0.334 to 0.262 and its DISTS
from 0.234 to 0.189. Wav2Lip moves the same way, and so does X-Dub when it is given the
target clip itself as its input instead of another clip of the person; X-Dub completes
51 of the 100 NeRSemble clips in this condition, and Table~\ref{tab:leak} compares the
two conditions on those clips. Part of the fidelity a paired evaluation credits to a
system run this way is therefore copied from its input.
MuseTalk-1.5 is run only in this form, so its row in Table~\ref{tab:main} carries the
same advantage.

\begin{table}[h]
\centering\footnotesize
\setlength{\tabcolsep}{5pt}
\renewcommand{\arraystretch}{0.9}
\caption{Same-frame reference leak. A \checkmark{} marks the rows whose reference or input
is the target clip.}
\label{tab:leak}
\begin{tabular}{@{}llccccc@{}}
\toprule
Set & Method (condition) & sees & MAE$\downarrow$ & LPIPS$\downarrow$ & DISTS$\downarrow$ & Tex$\to$1 \\
\midrule
\multirow{6}{*}{\rotatebox{90}{NeRSemble}}
& Wav2Lip (xref) & -- & 0.0463 & 0.473 & 0.340 & 0.200 \\
& Wav2Lip (official) & \checkmark & 0.0352 & 0.435 & 0.315 & 0.239 \\
& LatentSync-1.6 (xref) & -- & 0.0328 & 0.334 & 0.234 & 0.399 \\
& LatentSync-1.6 (official) & \checkmark & 0.0223 & 0.262 & 0.189 & 0.494 \\
& X-Dub (xvid, 51 clips) & -- & 0.0490 & 0.412 & 0.256 & 0.532 \\
& X-Dub (same-clip, 51 clips) & \checkmark & 0.0392 & 0.366 & 0.233 & 0.586 \\
\midrule
\multirow{4}{*}{\rotatebox{90}{HDTF}}
& Wav2Lip (xref) & -- & 0.0712 & 0.408 & 0.262 & 0.511 \\
& Wav2Lip (official) & \checkmark & 0.0446 & 0.322 & 0.219 & 0.571 \\
& LatentSync-1.6 (xref) & -- & 0.0490 & 0.280 & 0.183 & 0.605 \\
& LatentSync-1.6 (official) & \checkmark & 0.0325 & 0.191 & 0.135 & 0.700 \\
\bottomrule
\end{tabular}
\end{table}

\section{Implementation Details}\label{app:impl}
\paragraph{Networks of the objective.} The face parser of the shape term is a frozen
SegFormer \citep{p_segformer} fine-tuned on CelebAMask-HQ \citep{p_celebamaskhq}; its
mouth-interior and lip classes give the masks of Eq.~\ref{eq:dice}. The judge encodes a
grey, high-pass-filtered crop of the mouth with a ResNet-18 \citep{p_resnet},
pre-trained to tell people apart by their inner mouths, and compares the rendered mouth
with the reference through a small head on the two embeddings. The real mouths of other
people in the same training batch serve as its negatives.

\paragraph{Computational cost.} Training updates the 771.5M trainable parameters of
the parent generator and the HD-patch modules on four NVIDIA RTX 5090 GPUs (32\,GB each) in fp16 mixed
precision, with one 16-frame window per GPU per update. The reported model is trained
for 1{,}750 updates, with a peak of 21.2\,GiB of memory per GPU.

\section{Failure Cases}\label{app:failure}
Fig.~\ref{fig:failures} shows four failure cases of \ours{}, which the paragraphs
below explain.

\ifdefined\fcu\else\newlength{\fcu}\fi
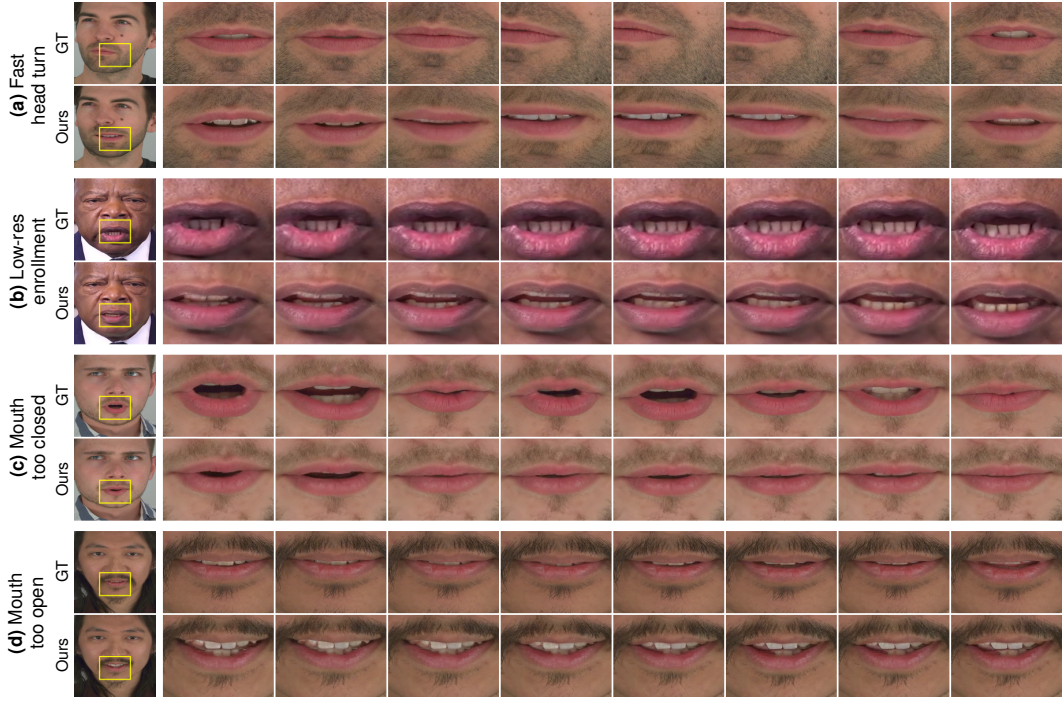
\begin{figure}[t]
\centering
\setlength{\fcu}{\dimexpr(\linewidth-26pt)/1711\relax}%
\begin{tikzpicture}[x=1\fcu,y=-1\fcu,font=\sffamily,
  fltag/.style={rotate=90,inner sep=0pt,font=\sffamily\tiny},
  flcase/.style={rotate=90,inner sep=0pt,align=center,font=\sffamily\scriptsize}]
\useasboundingbox ([xshift=-26pt]0,0) rectangle (1711,1202);
\node[inner sep=0pt,anchor=north west] at (0,0)
  {\includegraphics[width=1711\fcu]{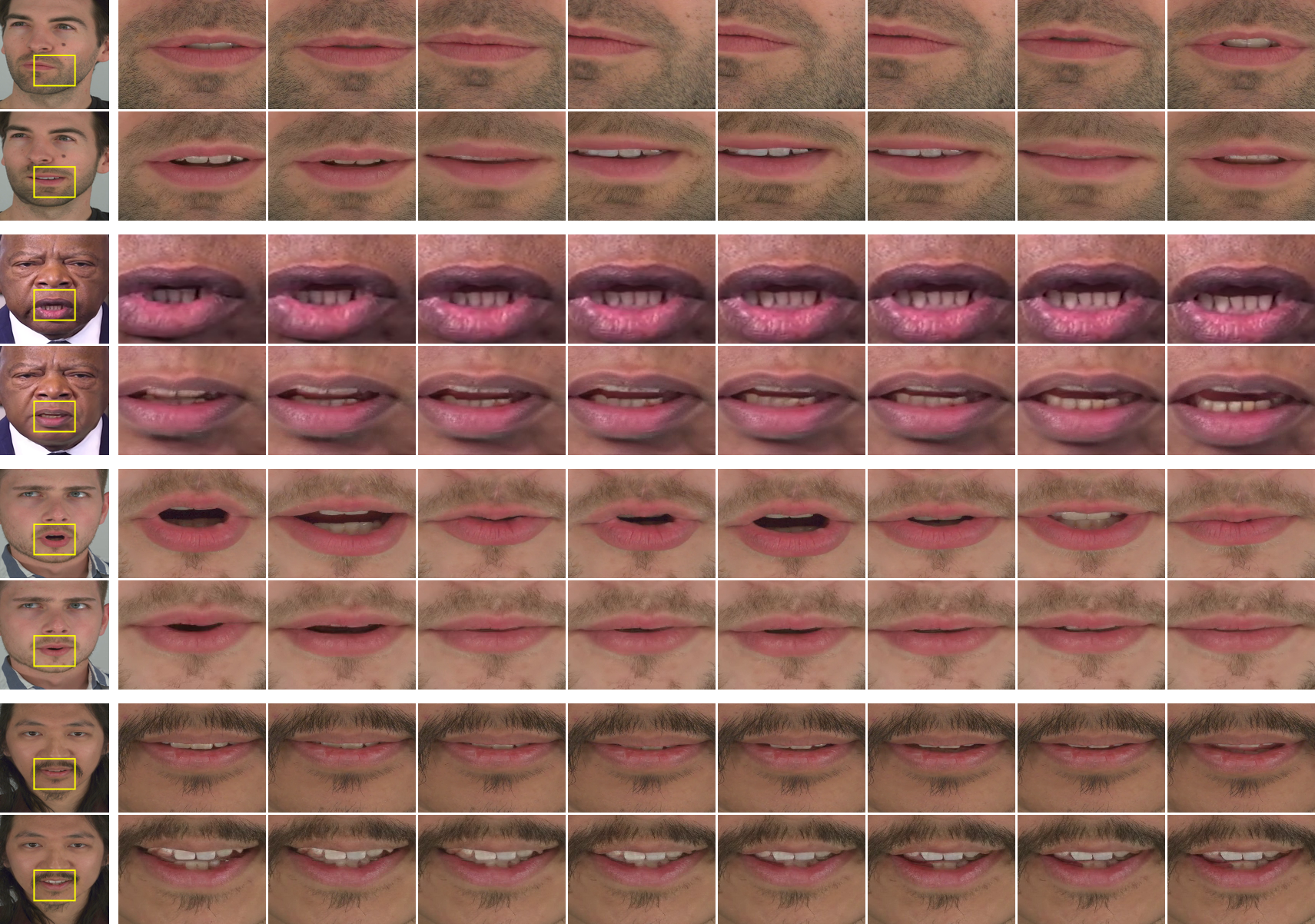}};
\foreach \k in {0,1,2,3}{%
  \node[fltag] at ([xshift=-4.5pt]0,305*\k+71) {GT};
  \node[fltag] at ([xshift=-4.5pt]0,305*\k+216) {Ours};}
\node[flcase] at ([xshift=-17.5pt]0,143.5) {\textbf{(a)} Fast\\head turn};
\node[flcase] at ([xshift=-17.5pt]0,448.5) {\textbf{(b)} Low-res\\enrollment};
\node[flcase] at ([xshift=-17.5pt]0,753.5) {\textbf{(c)} Mouth\\too closed};
\node[flcase] at ([xshift=-17.5pt]0,1058.5) {\textbf{(d)} Mouth\\too open};
\end{tikzpicture}
\caption{Failure cases of \ours{}: ground truth (top) and ours (bottom) at the same frames, beside the
full frame with its fixed oral box. (a) Fast head turn: our mouth stays frontal and is stretched across
the turned face. (b) Low-resolution enrollment (HDTF): the gapped lower teeth become a generic tooth row.
(c, d) Opening amplitude: the mouth is held too closed (c) or too open (d).}
\label{fig:failures}
\end{figure}

\paragraph{Fast head motion.} The judge and the fidelity metrics act on a fixed oral box in the aligned canvas. When the head turns quickly, the real
mouth turns with it toward the edge of that box, whereas \ours{} keeps a frontal mouth
inside the box, stretched across the turned face (Fig.~\ref{fig:failures}a); whatever
leaves the box is not seen by the judge.

\paragraph{Reference resolution bounds the result.} The method carries the person's
tooth and lip structure out of the enrollment references, so the finest detail it can
render is the detail those references contain. When the enrollment is itself
low-resolution in-the-wild video with a single reference clip, as on HDTF, there is no
fine structure to carry over and the mouth falls back to a generic one: in
Fig.~\ref{fig:failures}b the gapped lower teeth of the speaker become a generic tooth
row. We take this to be the main reason the advantage of Table~\ref{tab:main} does
not carry over to the out-of-distribution check of Appendix~\ref{app:hdtf}. The same
bound holds inside a studio recording for anatomy the enrollment never exposes:
sixteen enrollment frames cannot reveal a tooth surface the person never shows, the
judge then has nothing to compare against, and such frames revert to a plausible
generic mouth.

\paragraph{Opening amplitude follows the audio, not the person.} How wide the mouth
opens is predicted from the audio. Training matches the opening to the real frames on
average, but nothing tells the model how wide this particular person opens their mouth,
which affects every method we compare. The error goes both ways: the mouth can be held
too closed (Fig.~\ref{fig:failures}c) or opened too wide (Fig.~\ref{fig:failures}d),
and on average our open-area ratio is 0.880 on NeRSemble and 1.39 on HDTF. Making
opening amplitude an explicit conditioning variable, estimated for instance from the
person's own enrollment clips, is the natural next step.

Addressing these limitations, in particular controlling the opening amplitude and
judging a mouth that moves out of the oral box, remains an important direction for
our future work.

\subsection{Out-of-distribution check on HDTF}\label{app:hdtf}
HDTF \citep{zhang2021hdtf} is out of distribution for this setting on every axis the
method relies on: the source resolution is low with a small face region, the only
enrollment is a single other clip of the speaker, and head motion is larger. On it,
LatentSync-1.6 (xref) is better than \ours{} on the pixel and perceptual oral metrics
and on LSE-C and LSE-D (Table~\ref{tab:hdtf}). \ours{}, trained on NeRSemble studio recordings,
keeps the texture ratio closest to one and the lowest error outside the oral box, but
opens the mouth too wide.

\begin{table}[h]
\centering\scriptsize
\setlength{\tabcolsep}{4pt}
\renewcommand{\arraystretch}{0.9}
\caption{HDTF (48 query clips, 24 speakers) under the strict protocol; columns
and marks as in Table~\ref{tab:main}. Official Wav2Lip and LatentSync-1.6 rows are in
Table~\ref{tab:leak}. Ground truth's own SyncNet score is 7.72\,/\,7.05.}
\label{tab:hdtf}
\begin{tabular}{@{}lcccccccc@{}}
\toprule
Method (condition) & MAE$\downarrow$ & MAE-out$\downarrow$ & PSNR$\uparrow$ & LPIPS$\downarrow$ & DISTS$\downarrow$ & Tex$\to$1 & Open$\to$1 & LSE-C$\uparrow$/D$\downarrow$ \\
\midrule
Wav2Lip (xref) & 0.0712 & 0.0381 & 20.0 & 0.408 & 0.262 & 0.511 & 1.17 & 8.47/6.93 \\
LatentSync-1.6 (xref) & \textbf{0.0490} & 0.0199 & \textbf{22.7} & \textbf{0.280} & \textbf{0.183} & 0.605 & 0.711 & 9.30/5.70 \\
MuseTalk-1.5 (official)$^\dag$ & 0.0564 & 0.0161 & 21.6 & 0.348 & 0.239 & 0.647 & \textbf{0.918} & 7.38/7.33 \\
X-Dub (xvid) & 0.0991 & 0.0887 & 17.2 & 0.445 & 0.246 & \textbf{0.814} & 1.65 & 6.81/8.22 \\
sync.so lipsync-2 (xvid) & 0.0938 & 0.0859 & 17.7 & 0.435 & 0.235 & 0.801 & 1.10 & 1.73/13.0 \\
Kling lip-sync (xvid) & 0.0953 & 0.0872 & 17.5 & 0.425 & 0.235 & 0.760 & 1.29 & 3.82/10.6 \\
HeyGen lip-sync (xvid) & 0.0903 & 0.0839 & 17.9 & 0.418 & 0.224 & 0.665 & \textbf{1.08} & 8.24/6.49 \\
\midrule
\ours{} (ours) & 0.0605 & \textbf{0.0138} & 21.4 & 0.335 & 0.202 & \textbf{1.05} & 1.39 & 7.22/7.32 \\
\bottomrule
\end{tabular}
\end{table}

\end{document}